\documentclass[conference]{IEEEtran}
\IEEEoverridecommandlockouts  
\IEEEoverridecommandlockouts
\usepackage{booktabs}

\usepackage[utf8]{inputenc}

\usepackage{amsmath}
\usepackage[Symbol]{upgreek}

\usepackage{graphics}
\usepackage[tight]{subfigure}
\usepackage{hyperref}
\usepackage{bm}

\usepackage[]{footmisc}

\usepackage[ruled,vlined,linesnumbered]{algorithm2e}

\usepackage{amssymb}
\usepackage{amsfonts}

\usepackage{amsthm}
\usepackage{blindtext}
\usepackage{cite}
\usepackage{multirow}
\usepackage{scalerel}
\usepackage{tikz}

\usepackage{amsmath}

\usepackage[mathscr]{euscript}
\usepackage{dsfont}
\theoremstyle{definition}

\definecolor{orcidlogocol}{HTML}{A6CE39}
\tikzset{
  orcidlogo/.pic={
    \fill[orcidlogocol] svg{M256,128c0,70.7-57.3,128-128,128C57.3,256,0,198.7,0,128C0,57.3,57.3,0,128,0C198.7,0,256,57.3,256,128z};
    \fill[white] svg{M86.3,186.2H70.9V79.1h15.4v48.4V186.2z}
                 svg{M108.9,79.1h41.6c39.6,0,57,28.3,57,53.6c0,27.5-21.5,53.6-56.8,53.6h-41.8V79.1z M124.3,172.4h24.5c34.9,0,42.9-26.5,42.9-39.7c0-21.5-13.7-39.7-43.7-39.7h-23.7V172.4z}
                 svg{M88.7,56.8c0,5.5-4.5,10.1-10.1,10.1c-5.6,0-10.1-4.6-10.1-10.1c0-5.6,4.5-10.1,10.1-10.1C84.2,46.7,88.7,51.3,88.7,56.8z};
  }
}

\newcommand\orcidicon[1]{\href{https://orcid.org/#1}{\mbox{\scalerel*{
\begin{tikzpicture}[yscale=-1,transform shape]
\pic{orcidlogo};
\end{tikzpicture}
}{|}}}}
\usepackage{hyperref}

\usepackage{eso-pic}

\graphicspath{{./figs/}}

\title{
Graph Attention Multi-Agent Fleet Autonomy \\
for Advanced Air Mobility
}
\author{Malintha Fernando, Ransalu Senanayake, Heeyoul Choi, Martin Swany 
\thanks{ Malintha Fernando, Heeyoul Choi, and Martin Swany are with the Luddy School of Informatics, Computing, and Engineering  at Indiana University, Bloomngton, IN, 47401, USA. E-mail:{\tt\small ccfernan@iu.edu, hchoi@handong.edu, swany@iu.edu}} 
\thanks{Ransalu Senanayake is with Stanford University, CA, 94305, USA. E-mail:{\tt\small ransalu@stanford.edu}.}
\thanks{This research was partly funded by the National Institute of Standards and Technology (NIST) through the grant 70NANB21H037.}
\thanks{Heeyoul Choi would like to acknowledge the support from IITP grant of Korea (No. 2018-0-00749) and Basic Science Research Program through NRF of Korea (2022R1A2C1012633).}
\thanks{We also thank Lilly Endowment, Inc., for its support to Indiana University Pervasive Technology Institute.}
}

\newcommand{\rebutt}[1]{\textcolor{black}{{}#1}}

\begin{document}

\maketitle

\begin{abstract}
Autonomous mobility is emerging as a new disruptive mode of urban transportation for moving cargo and passengers. 
However, designing scalable autonomous fleet coordination schemes to accommodate fast-growing mobility systems is challenging primarily due to the increasing heterogeneity of the fleets, time-varying demand patterns, service area expansions, and communication limitations.
We introduce the concept of partially observable advanced air mobility games to coordinate a fleet of aerial vehicles by accounting for the heterogeneity of the interacting agents and the self-interested nature inherent to commercial mobility fleets. To model the complex interactions among the agents and the observation uncertainty in the mobility networks, we propose a novel heterogeneous graph attention encoder-decoder (HetGAT Enc-Dec) neural network-based stochastic policy. We train the policy by leveraging deep multi-agent reinforcement learning, allowing decentralized decision-making for the agents using their local observations. Through extensive experimentation, we show that the learned policy generalizes to various fleet compositions, demand patterns, and observation topologies. Further, fleets operating under the HetGAT Enc-Dec policy outperform other state-of-the-art graph neural network policies by achieving the highest fleet reward and fulfillment ratios in on-demand mobility networks.
\end{abstract}
\section{Introduction}
The latest advancements in aerial robotics and battery technologies are paving the way for \textit{Advanced Air Mobility (AAM)}: a new disruptive mode of transportation that focuses on moving cargo and passengers using electric-powered Unmanned Aerial Vehicles (UAV) operating at low altitudes over short distances   \cite{bradford2020concept}.
With an appealing node-to-node navigation structure that overreaches 
already exhausted and poorly maintained path-based ground transportation networks, AAM is currently emerging as a sustainable and efficient alternative to solve the \textit{last-mile delivery problem} in retail and logistics sectors \cite{vasani_2022}.

Thanks to their vast operational space, superior maneuverability, relative affordability, efficiency, and state-of-the-art collision-avoiding capabilities, the AAM fleets face lesser risks in scaling than their ground-based counterparts across a wide range of novel commercial applications.
However, the existing centrally controlled air traffic systems are significantly limited in their ability to cater to the rapidly growing UAV market, thus onboard, autonomous decision-making approaches appeal increasingly for coordinating UAV fleets in a decentralized manner \cite{bradford2020concept, lappas2022eurodrone}.
Delegating the high-level decision-making in real-world commercial AAM applications yet poses numerous challenges; the \textit{dynamic} fleet sizes, stochastic communication, maximizing returns in \textit{high owner-to-vehicle affinity}, and heterogeneity within the mobility networks, to name a few.
\begin{figure}[t!]
    \centering
    \includegraphics[trim={0cm 0cm 0cm 0cm}, clip, width=0.49\textwidth]{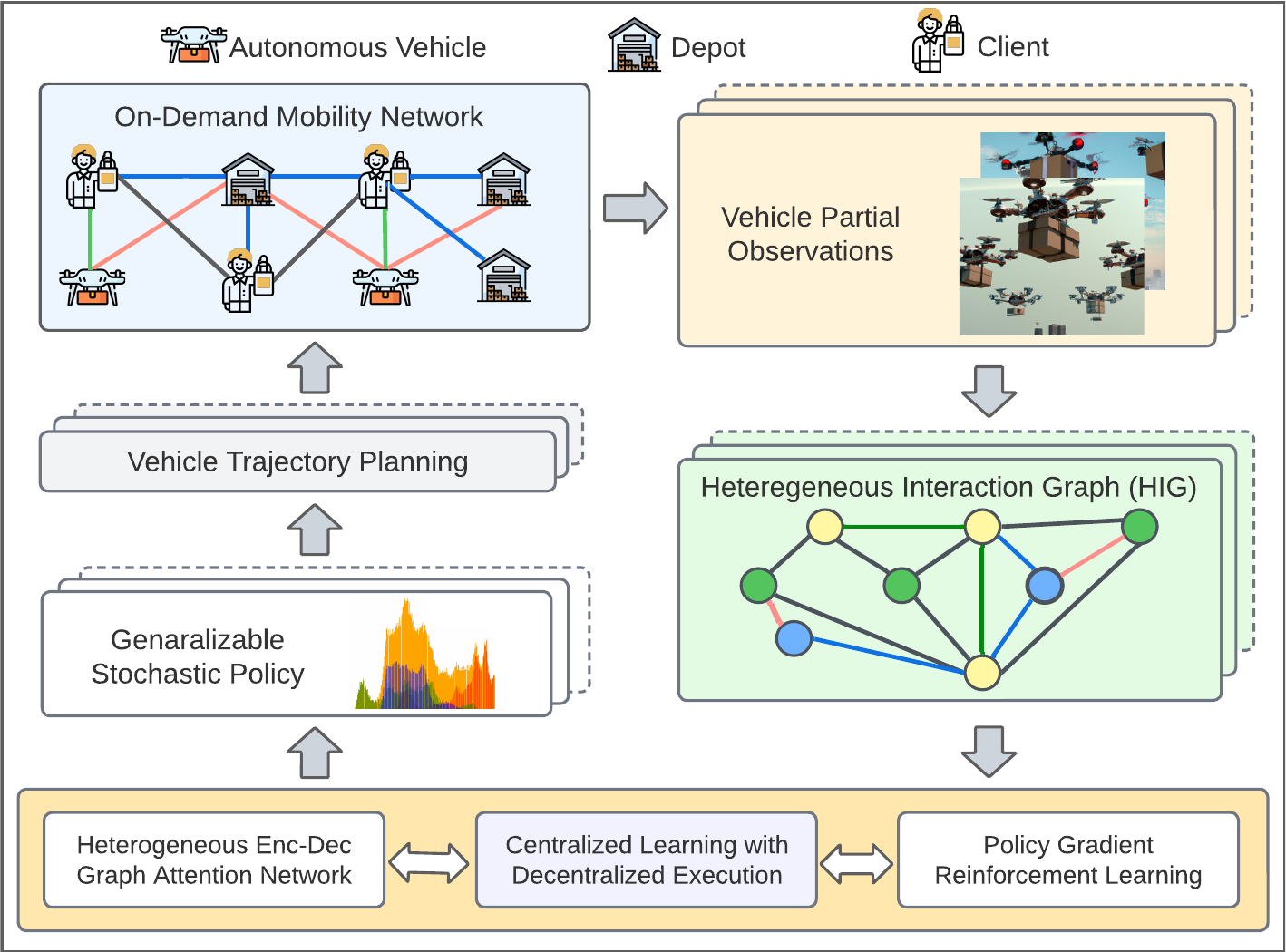}
    \caption{The overview of the presented approach. The on-demand mobility network has multiple service-providing depots, UAV agents, and clients corresponding to three \textit{meta-types}. The red, green, and black lines represent interactions between the each meta-type entity.
    The nodes and edges of the local Heterogeneous Interaction Graph (HIG) correspond to interacting entities and their semantic relations within one's observation range.
    We represent the each meta-type node in yellow, green, and blue colors. 
    The Heterogeneous Graph Attention Encoder-Decoder (HetGAT Enc-Dec) neural network directly operates on the HIG to compute a stochastic policy for decentralized decision-making.
    Drones flying digital arts: DALL-E, OpenAI \copyright.
    }
    \label{fig:cover}
\end{figure}

We formulate AAM as a partially observable stochastic game (POSG) that is inherently decentralized and enables \textit{self-interested} autonomous vehicle agents to make high-level decisions by accounting for the time-varying heterogeneous interactions within a complex mobility system. 
Fig. \ref{fig:cover} shows the interactions among different entities in a mobility network considered in this work; a heterogeneous UAV fleet, depots populating various payloads, and destination clients.
Compared to existing work, where the mobile agents are fully-cooperative \cite{tsao2019model} or coordinated by a central policy \cite{gammelli2021graph}, the game-theoretic formulation captures the revenue-seeking behavior of the high-affinity commercial fleets, i.e., taxi fleets.
This formulation notably leads to a more practical \textit{general-sum} game, where the agents' rewards are related arbitrarily, thus, mixed cooperative-competitive in nature, as opposed to fully-cooperative or fully-competitive games with either shared or zero-sum reward functions.
The general-sum nature of the AAM game helps us using a practical reward computing method for the autonomous vehicle agents building on the non-linear taxi-fare calculation proposed in \cite{yang2010nonlinear}.

We argue that explicit model-based solutions are not ideal to solve the AAM game due to 1) the inherent difficulty of solving general-sum games, 2) the partial observability caused by wireless communication limitations that prevent the autonomous vehicles from aggregating the \textit{full} fleet state \cite{fernando2022coco}, 3) and the fleet heterogeneity.
Thus, we propose a deep multi-agent reinforcement learning (MARL) approach for coordinating vehicle agents in a heterogeneous fleet by accounting for their interactions.
Specifically, we build on the premise that an agent's heterogeneous interaction topology constitutes a graph corresponding to a set of \textit{meta-type} nodes and edges, whose relations are quantifiable by a \textit{Heterogeneous Graph Neural Network} (HetGNN) to learn a generalizable policy.

By following the notion of graph attention, we propose a novel heterogeneous graph attention encoder-decoder (HetGAT Enc-Dec) policy where the \textit{attention} mechanism computes a score between the interacting node features \cite{kool2018attention, vaswani2017attention}.
To handle the \textit{non-stationarity} issue in MARL caused by the ever-changing policies of the other agents, we leverage \textit{centralized training and decentralized execution} (CTDE) paradigm \cite{oroojlooy2022review}.
The CTDE MARL allows the agents to access the experiences collected by the other agents during the training by sharing the policy parameters, yet requiring only local observations for the decision-making.
Fig. \ref{fig:cover} shows the deep MARL training and the decision-making loops for an autonomous vehicle by using a heterogeneous graph, constructed using the local observations.

Through extensive experimentation, we show that the proposed approach is highly generalizable to varying fleets, environments, demand patterns, and observational topologies, thus rendering it suitable for coordinating autonomous vehicle fleets in dynamic and complex mobility environments.
We additionally introduce an \textit{intrinsic} fleet rebalancing mask based on a vehicle's local observations that improve the policy's performances under varying demand patterns.
The main contributions of this work are,
\begin{itemize}
    \item formulating AAM as a POSG for coordinating vehicle agents with \textit{hierarchical-timescale} autonomy by accounting for complex, heterogeneous interactions in mobility networks, and the general-sum fair calculation in transportation literature (Section IV),
    \item proposing a novel HetGAT Enc-Dec architecture for autonomous mobility under time-varying partial observation topologies, demand-patterns by performing intrinsic fleet rebalancing (Section V),
    \item \rebutt{evaluating the deep MARL solution performances of the stochastic AAM game against different Graph Neural Network (GNN) policy architectures by drawing connection to the \textit{social optimum}, and the agents' observation topology (Section VI)}.
\end{itemize}
To the best of the authors' knowledge, HetGAT-based MARL has not yet been studied in the on-demand mobility context under the POSG constraints considered in this paper, thus making our work the first of its kind.

\section{Related Work}

\subsection{\rebutt{Neural Networks for Learning Graph-Structured Data}}
\rebutt{
A plethora of real-world data takes the form of graphs, e.g., social and computer networks, protein interactions, etc. 
The GNNs have emerged as a powerful tool for learning from such data by extending the traditional neural network architectures to operate directly on the input graph-structured data, to capture complex relationships and dependencies among the nodes \cite{zhou2020graph}.
Compared to conventional convolution neural networks, GNNs share the convolution operators across the graph; thus, generalizable to graphs of various sizes and degrees \cite{kipf2016semi, gori2005new}.
}
Graph attention neural networks (GAT) advance GNN by prioritizing the neighboring node features by through an attention score before \textit{aggregating} together, depending on the features' prominence toward the learning task \cite{velivckovic2017graph}.
In \cite{kool2018attention}, authors presented a GAT method for sequential routing plans proving their robustness and generalizability in the combinatorial optimization domain.

The Heterogeneous variant of GNN improves the model interpretability and expressiveness by allowing more complex graph structures with multiple heterogeneous relations among different meta-type entities.
In contrast to homogeneous GNN, the convolution operators in HetGNN are type-specific and thus operable on varying feature spaces.
Following the success of GNN, the HetGAT show promise in parallel research directions for learning generalizable policies in combinatorial and sequential decision-making tasks: multi-robot task allocation \cite{wang2022heterogeneous}, sequential traffic speed prediction \cite{jin2021hetgat}.
In \cite{li2021message} authors propose a large-scale multi-agent path planning with GATs for attentive bandwidth consumption in limited communication settings.

\subsection{\rebutt{Deep Multi-Agent Reinforcement Learning}}

In contrast to fully-cooperative or fully-competitive games, which have been often discussed in the stochastic games literature, solving general-sum POSG for a stationary \textit{Nash equilibria}, especially in multi-agent settings, remains an open challenge \cite{ yang2020overview, zhang2021multi}.
The state-of-the-art DRL approaches have contributed significantly to the recent advancements related to computing multi-agent coordination policies in POSG environments; e.g., in \cite{vinyals2019grandmaster}, authors achieved superhuman performances in StarCraft II.
In a parallel research direction, Lowe et al. \cite{lowe2017multi} showed that actor-critic algorithms trained using CTDE to generate robust multi-agent policies in POSG environments with complex inter-agent relationships.

The GNN-based MARL methods range in their rewarding mechanism across numerous application domains.
In \cite{seraj2022learning,gupta2017cooperative}, authors consider fully-cooperative decision-making for multi-agent teams, where the latter operates in heterogeneous settings.
In \cite{deka2021natural}, authors study the naturally emerging behavior in partially observable multi-agent team games in the presence of two types of interactions occurring among the agents.
In \cite{zhang2022multi}, authors discuss a HetGAT-based multi-agent approach for training an electric vehicle charging pricing policy.
Our approach, however, stretches beyond the existing works to incorporate complex, topological interactions occurring among multiple entity types, along with heterogeneity within themselves.

\subsection{Autonomous Mobility Fleet Coordination}

Current autonomous mobility fleet coordination spans multiple research areas; autonomous mobility-on-demand (AMoD) \cite{carron2019scalable}, multi-robot dynamic task allocation \cite{choudhury2022dynamic}, drone-assisted delivery \cite{oh2018task} and robot pick up and delivery systems \cite{salzman2020research}.
Many AMoD solutions consider a centralized policy that coordinates the vehicles with single occupancy \cite{gueriau2020shared, gammelli2021graph} or ridesharing \cite{wallar2018vehicle, meneses2022optimization}.
However, the latter notion is still far from AAM due to unique safety and infrastructural regulations.
In \cite{gammelli2021graph}, Gammelli et al. present a Graph Neural Network (GNN)-based centralized policy that is also generalizable to different service areas for coordinating an AMoD fleet.  
The autonomous mobility fleet redistribution under congestion has also been studied with Q-learning  \cite{gueriau2020shared} and optimization-based \cite{solovey2019scalable} approaches.
Especially, Gu{\'e}riau et al. \cite{gueriau2020shared} propose simultaneous pick up, delivery, and rebalancing using RL agents with fleet \textit{elasticity}. 
The agents' action spaces, however, limit to the closest ride requests, whereas ours are free to choose any depot to encourage exploration, and for implicit fleet rebalancing.
The model predictive control (MPC) methods have also been used to solve AMoD; with a composite, weighted utility function to maximize the fleets' and the riders' rewards\cite{tsao2019model}. 
Carron et al. \cite{carron2019scalable} propose an MPC-based method for AMoD with explicit system delay modeling, yet overlooks the vehicles and payloads with various capacities.

Multi-agent pick up and delivery \cite{xu2022multi} recently received a spotlight as a viable direction for coordinating warehouse and mobility fleets.
In \cite{liu2019task, choudhury2021efficient}, authors propose a hybrid approach for the multi-agent pick up and delivery problem that simultaneously addresses the path planning.
The latter work combines drone package delivery with public transit systems to conserve energy.
Choi et al. \cite{choi2017optimization} propose a drone multi-package delivery focusing on battery and payload constraints, albeit overlooking the on-demand perspective.
In \cite{alkouz2021reinforcement}, authors propose a drone swarm redistribution approach using a centralized policy.
Although our approach does not perform explicit multi-agent path planning, it supports training agents with \textit{multiple timescale} autonomy systems to account for path planning, low-level control, and trajectory optimization hierarchically.

Dynamic task allocation (DTA) introduces temporal constraints to otherwise spatially-constrained conventional task allocation algorithms.
In \cite{choudhury2022dynamic, wang2022heterogeneous}, authors propose multi-robot dynamic task allocation approaches, with the former considering a drone package delivery task under temporal uncertainty.
However, the oversight of robots' movements makes them better suited for in-place task completion over mobility applications.

\section{Background}

\subsection{Partially Observable Stochastic Games}
Stochastic games extend the Markov decision processes (MDP) to the multiple agents setting \cite{littmanMarkovGamesFramework1994}, where the stochasticity stems from the simultaneous action selection of the agents.
Partially observable stochastic games are most suited for environments where the system state is not fully visible to the agents due to practical limitations; consequently, they must make decisions using individual local observations.
It mainly differs from the seemingly similar decentralized partially observable Markov decision processes (Dec-POMDP) by allowing the agents to act in their \textit{self-interest}; whereas the agents in the latter share identical reward functions \cite{yang2020overview}.

We define a POSG as an eight-tuple $\langle N$, $S$, $\mathbb{T}$, $ \{{R}_{i \in 1,\dots, N}\}$, $\{A_{i\in 1,\dots, N}\}$, $\gamma$, $\{O_{i \in 1,\dots, N}\}$, $\mathbb{O}$$\rangle$, where $N$ is the number of agents in the game, $S$ is the full state space,  $A_i$ and $O_v$ are the action and the local observation spaces for agent $i$. 
For a given action profile $\mathbb{\textbf{A}} = a_1 \times \dots \times a_N$, $\forall a_i \in A_i$, the state $S$ changes according to the state transition function $\mathbb{T}: S \times \mathbb{\textbf{A}} \mapsto S'$, and agent $i$ receives a reward defined by the function $R_i: S \times A_i \mapsto \mathbb{R}$ according to its action.
Here $\gamma \in [0,1]$ is a discount factor.
\label{def:posg}
\begin{figure*}[t]
    \centering
    \subfigure[]
    {\includegraphics[trim={0cm 0cm 0cm 0cm}, clip, width=0.23\textwidth]{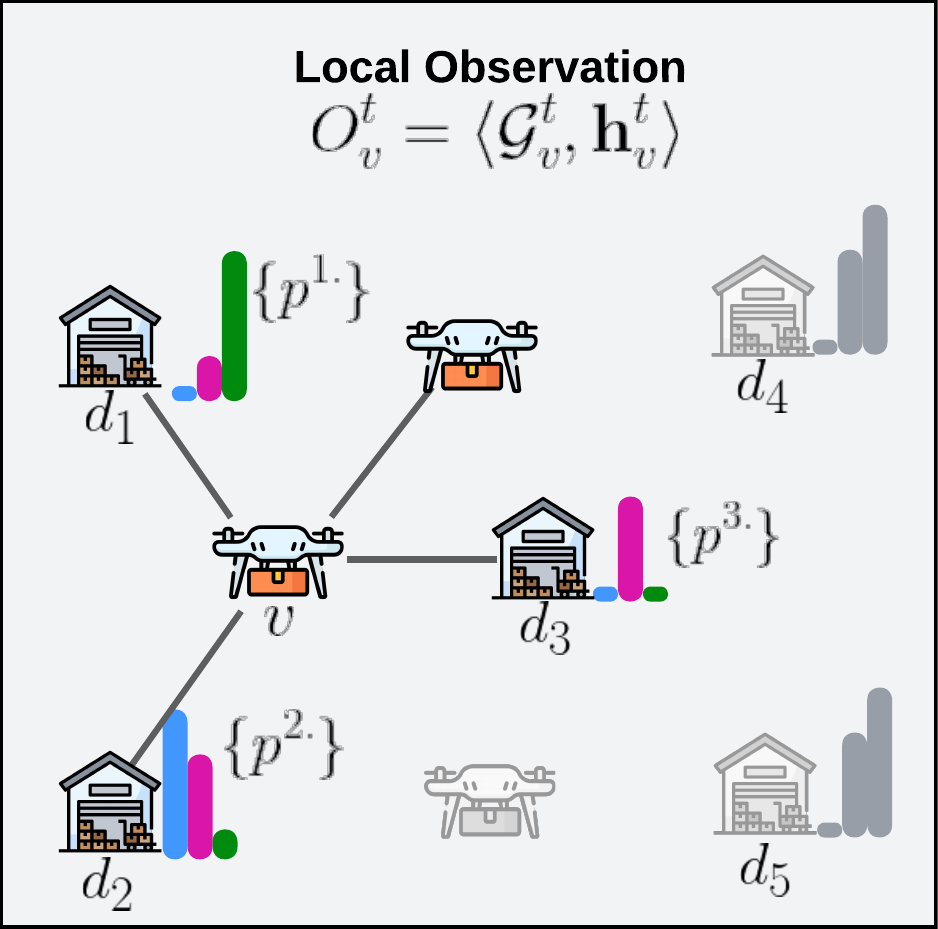}
    \label{fig:obser}}
    \subfigure[]
    {\includegraphics[trim={0cm 0cm 0cm 0cm}, clip, width=0.23\textwidth]{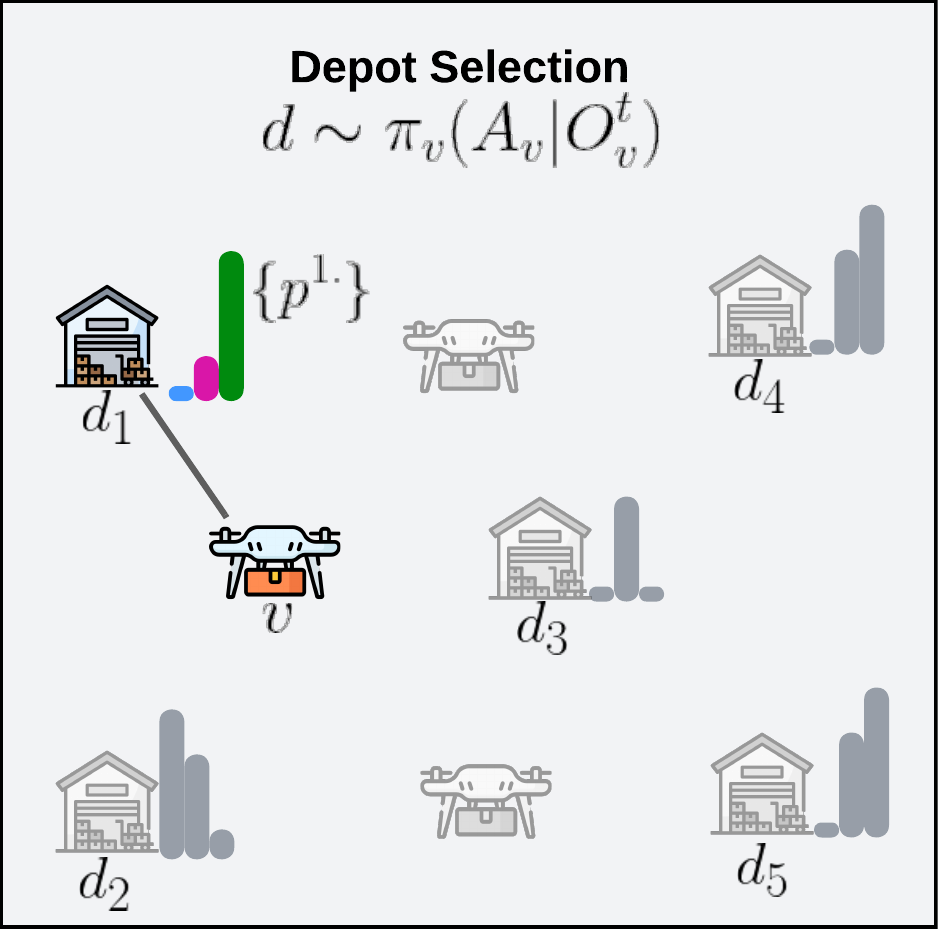}
    \label{fig:selection}}
    \subfigure[]
    {\includegraphics[trim={0cm 0cm 0cm 0cm}, clip, width=0.23\textwidth]{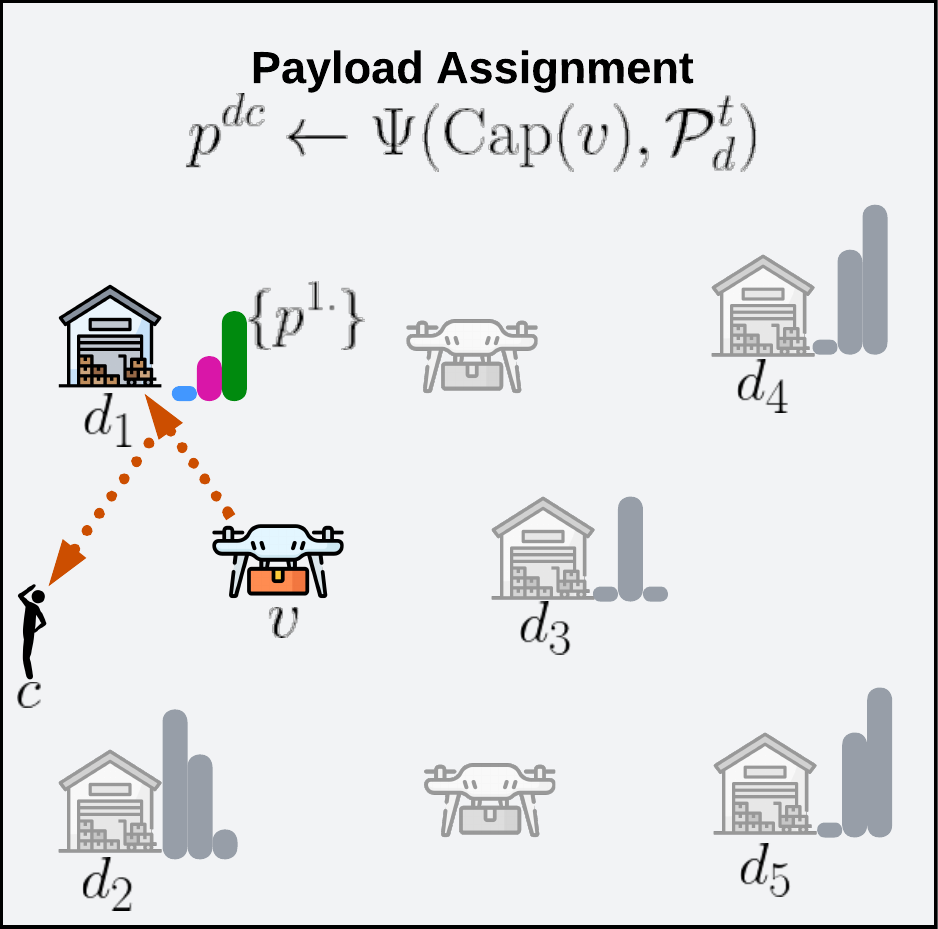}
    \label{fig:assignment}}
    \subfigure[]
    {\includegraphics[trim={0cm 0cm 0cm 0cm}, clip, width=0.23\textwidth]{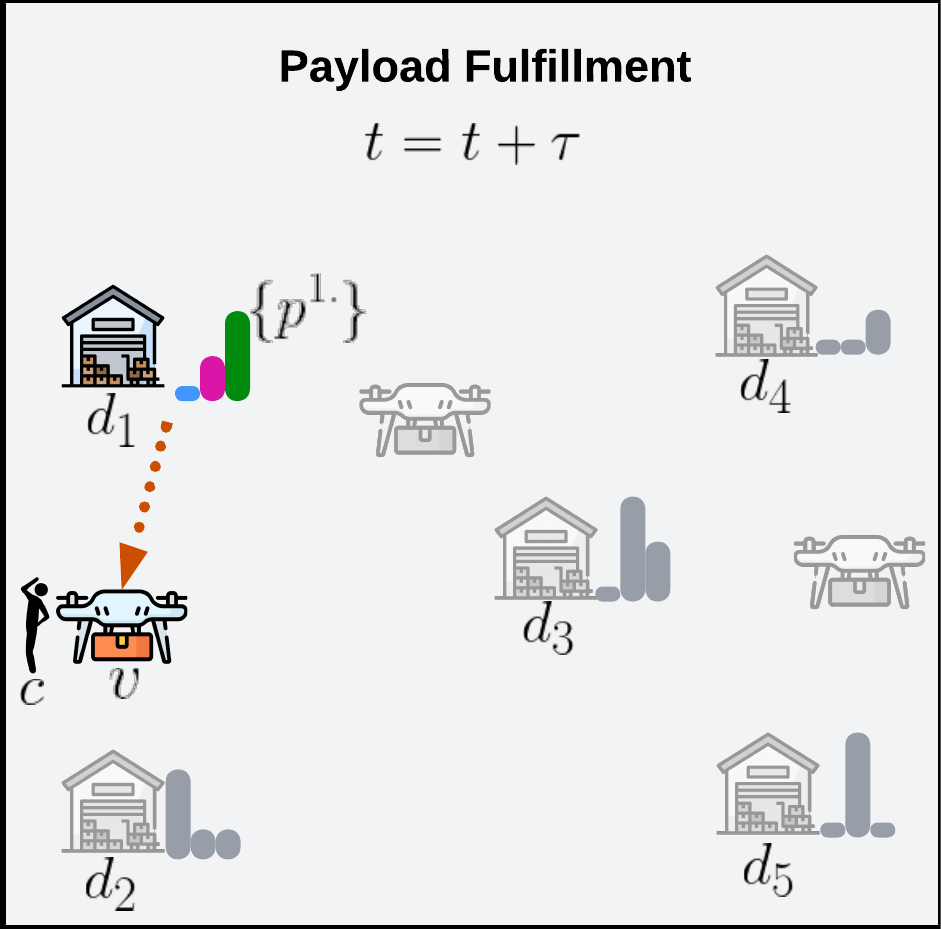}
    \label{fig:fulfillment}}
    
    \caption{Different stages of payload fulfillment by a single UAV agent. \textbf{(a)} The local observation space of agent $v$ at time $t$. The color images and solid black lines show agents' observable neighbors, and their communication links. The blue, magenta and green color bars denote each type of payload at the depots. \textbf{(b)} Agent $v$ selects a depot using its policy $\pi_v$ using the observations and communicates its selection. \textbf{(c)} The depot assigns a payload to the agent from its available payload requests set. Note the amount of green color payloads is reducing. \textbf{(d)} Agent fulfills the payload request by traveling to the chosen depot and next to the assigned client $c$. Here $\tau = \tau_1 + \tau_2$ denotes the total travel time. }
    \label{fig:fullfilment diagram}
\end{figure*}
The observation function maps the state $S$ to the agents' local observations $\mathbb{O}: S \mapsto O_i$.
The objective of a POSG is to find an optimal policy $\pi_i$ which maximizes the  agent $i$'s \textit{expected cumulative discounted reward}, $J(\pi_i) = \mathbb{E}_{a_i \sim{\pi_i}}[\sum_t^\mathbf{T} R_i (s^t, a^t_i)]$ using their local observations.
In this work, we consider \textit{general-sum} rewards calculated according to the trip distance and the payload size using a non-linear taxi-fair calculation method following \cite{yang2010nonlinear} (see \ref{sec:reward} for more on the reward calculation). 

\subsection{Policy Gradient Deep Reinforcement Learning}
DRL computes an optimal policy $\pi_{\theta_i}(a^t_i| o^t_i)$ characterized by a set of parameters $\theta$ using the agents' experiences acquired during the training.
As the agent's action space increases, the exploration step of many action-value methods, e.g., Q-Learning, becomes intractable, leading to impractical training times.
The policy gradient (PG) methods particularly excels in tasks involving large state and action spaces as it directly optimizes the policy parameters $\theta$ in the direction of the policy gradient \cite{sutton1999policy, silver2014deterministic}.
Let $J(\pi_{\theta_i})$ defines the vanilla policy gradient,
\begin{equation}
J(\pi_{\theta_i}) = \hat{\mathbb{E}}_{a_i \sim{\pi_{\theta_i}}} \large[ \triangledown_{\theta_i} \log \pi_{\theta_i}(a_i|O_v) Q^{\pi_i}(S,a_i) \large],
\label{eq:policy_gradient}
\end{equation}
where $\hat{\mathbb{E}}_{a_i \sim{\pi_{\theta_i}}}(.)$ and $Q^{\pi_i}(S,a_i)$ are empirical expectation and the action-value function for agent $i$.
In this work, we use proximal policy optimization (PPO) for training the stochastic policy that replaces  $Q^{\pi_i}(S,a_i)$ in Eq. \ref{eq:policy_gradient} using a \textit{clipped surrogate objective} and an \textit{advantage estimator} \cite{schulman2017proximal}.
The clipped surrogate objective prevents PPO from updating the policy parameters too aggressively, thus stabilizes the learning including in multi-agent settings \cite{yu2021surprising}.
In our experiments, we observed PPO to further generate higher fleet rewards compared to Q-Learning and vanilla policy gradient methods.
We further use actor-critic DRL for training the policy, where a \textit{critic} network provides \textit{value} estimations to the policy, also known as the \textit{actor}.










\section{Partially Observable AAM Game}

\subsection{Hierarchical Timescales}
Autonomous vehicle agents operating in a mobility system require specific tailoring as opposed to many other RL agents; mainly due to their 1) hierarchical action execution and 2) asynchronous action selection nature.
For instance, two vehicles might not complete their journeys simultaneously: one may need to choose a new depot to undertake a payload while the other is still traveling.
This stems from the hierarchical timescale nature of the vehicle autonomy where the execution of a high-level decision relies on multiple low-level components operating at different frequencies, such as motion planning and control \cite{chung2018survey}.
As a result, choosing an action by a vehicle agent intermediary, while committed to a travel may disrupt its current trajectory, causing unnecessary observations to aggregate in the experience buffer affecting the training results and the performances negatively.
Simultaneously, it is also crucial to update one's internal state throughout the simulation, so that the other agents can observe its state.
We believe such real-world constraints must be accounted for in designing MARL-based robotic simulation frameworks to maximize training performances.
In this work, we propose a hierarchical timescale approach by introducing the notion of \textit{active timesteps}.

The mobility network evolves in small, discrete timesteps $\Delta t$.
Consider an identity function that indicates a vehicle $v$'s availability at time $t$ such that $\mathds{1}_{avail}(v^t) = 1$ is when $v$ is available to undertake payloads, or $\mathds{1}_{avail}(v^t) = 0$ is when it is committed to a payload request and unavailable.
We only consider a timestep $t$ as an \textit{active timestep} if it results in a change of the vehicle's availability function such that,
a timestep $t$ is active iff $\mathds{1}_{avail}(v^t) \neq \mathds{1}_{avail}(v^{t - \Delta t})$.
Throughout this work, we consider the vehicles' action selections and observations only occur at active timesteps, leaving the local trajectory execution and UAV control to take place intermediary.


\subsection{Partially Observable Stochastic AAM}
Let $\mathcal{D}$, $\mathcal{C}$ and $\mathcal{V}$ denote a set of stationary \textit{depots}, \textit{clients} and a fleet of heterogeneous UAVs.
The depots may resemble warehouses or designated locations in a mobility network where the robots can pick up payloads that need to be delivered to another depot or a client location. 
For brevity, we refer to a destination as a client, and denote by $c$.
Let $x_{v}^t, x_{d}, x_{c} \in \mathbb{R}^2$ denote the locations of a vehicle $v \in \mathcal{V}$, a depot $d \in \mathcal{D}$ and a client $c \in \mathcal{C} \cup \mathcal{D} \setminus d$.
Let $p^{dc} \in \mathcal{P}_d$ denote a \textit{payload request} indicating a payload located at depot $d \in \mathcal{D}$ with a destination $c$, and $\mathcal{P}_d^t$ is the state of the \textit{payload queue} at $d$ at time $t$.
At a given timestep $t$, the system may contain an arbitrary number of payload requests in each queue.
Each UAV in the system may communicate with its neighboring UAVs and the depots to acquire its local observations as shown in Fig. \ref{fig:obser}.
Thus, we define a time-varying neighborhood for a vehicle ${v}$, $\mathcal{N}_{v}^t$ containing its observable vehicles $\mathcal{V}_v^t$ and the depots $\mathcal{D}_v^t$ at the active timestep $t$, such that $\mathcal{N}_{v}^t \in \mathcal{D} \cup \mathcal{V}$.
Complementing the structure of the graph input data in GNN literature, we define vehicle observations as a tuple $O_v$ $=$ $\langle \mathcal{G}_v^t, \mathbf{h}_v^t \rangle$, where $\mathcal{G}_v^t$ is a time-varying heterogeneous interaction graph (HIG) constitutes to the topology of the neighborhood $\mathcal{N}_{v}^t$.
Specifically, the nodes and the edges of a HIG corresponds to the elements of $\mathcal{N}_{v}^t$, and the presence of an interactions between any two nodes.
The feature space of an observation is denoted by $\mathbf{h}_v^t$. 

First, a vehicle $v \in \mathcal{V}$ where $\mathds{1}_{avail}(v^t) = 1$ chooses a depot $d \in \mathcal{D}$  given its local observations as shown in Fig. \ref{fig:selection}, and communicates the selection to $d$.
Let $\mathrm{Cap}(v)$ define the maximum capacity of vehicle $v$.
We categorize the payloads by size, such that a vehicle with capacity $\mathrm{Cap}(v)$ can only fulfill payload requests of size $\mathrm{Cap}(p^{.})$, where $\mathrm{Cap}(p^{.}) \leq \mathrm{Cap}(v)$. 
The depot assigns the agent a payload $p^{dc}$ from its payload queue $\mathcal{P}_d^t$ using a fixed assignment function $\Psi$ considering the vehicle's maximum capacity, such that $\Psi: \mathrm{Cap}(v) \times \mathcal{P}^t_d \mapsto p^{dc}$ for $p^{dc} \in \mathcal{P}^t_d$.
The depot next discard the payload request from the queue  $\mathcal{P}_d^{t+\Delta t} = \mathcal{P}_l^{t} \setminus p^{dc}$ as in Fig. \ref{fig:assignment}.

Upon the assignment, $v$ switches to unavailable mode $\mathds{1}(v^{t + \Delta t}) = 0$, visits the chosen depot $d$ to pick up the payload.
Finally, it travels to the destination $c$ to drop off the payload, at which point it switches back to $\mathds{1}(v^{t+\tau + \Delta t}) = 1$ as shown in Fig.\ref{fig:fulfillment}.
Let $\tau_1$, $\tau_2$ denote the time it takes for $v$ to travel to the $d$ from the current location, and to $c$ from $d$. Let $\tau = \tau_1 + \tau_2$.
Upon completing the travel at $ t+\tau$, $v$ collects a \textit{net} reward computed from the \textit{payoff} specified in the payload request and the vehicle's initial state.
If the chosen depot $d$ does not contain a suitable payload for $v$, the vehicle may stop at $d$, collects a negative net reward, and marks itself as available $\mathds{1}_{avial}(v^{t+\tau_1 + \Delta t}) = 1$.

We formally define each vehicle $v$'s objective as,
\begin{subequations}
\begin{alignat}{3}
\mathrm{Maximize}  \quad & \mathbb{E}_{a_v \sim \pi_v} \Big [ \sum_{t = 0}^{T} \mathds{1}_{avail}(v^t) \gamma^t         
    R_v(s^t, a_v^t) \Big ], 
    \label{eq:obj} \\        
     \mathrm{Subject \, to}  \quad & a^t_v \sim \pi_v(A_v | O_v^t), A_v = \mathcal{D},
     \label{eq:cons1}
    \\
    & O^t_v = \langle \mathcal{G}_v^t, \mathbf{h}_v^t \rangle, \\
    & O^{t+\tau}_v = \langle \mathcal{G}_v^{t+\tau}, \mathbf{h}^{t+\tau}_v \rangle \\
    &  p^{dc} \xleftarrow{} \Psi(\mathrm{Cap}(v), \mathcal{P}^t_d) \\
    & \mathrm{Cap}(v) \leq \mathrm{Cap} (p^{dc}) \\
    & r_v \xleftarrow{} R_v: x_{v}^t \times h_{p^{dc}}, \\
    & \mathds{1}_{avail}(v^{t}) = \mathds{1}_{avail}(v^{t + \tau_1 + \tau_2}) = 1, 
    \label{eq:cons_reward}
  \end{alignat}
  \label{eq:objective}
\end{subequations}
\noindent where
$t+\tau \leq T$ is the planning horizon, $h_{p^{dc}}$ is the features associated with the payload request.
We seek a generalizable stochastic policy that is shareable by all the vehicle types in the fleet for making local decisions that maximize their rewards.
Additionally, the resulting policy must scale to a variety of heterogeneous depots and vehicles in the fleet to accommodate dynamic addition and removal of entities.
Thus, it must suit for coordinating fleets elastic size conditions, and support service area expansion mimicking the requirements of real-world mobility applications.

\section{Graph Attention MARL For Solving AAM }

We start by constructing the HIG that subsumes the interactions among different meta-type entities in the mobility network.
Each edge in the HIG represents a specific relation between two interacting entities, and belongs to a set of semantic relations we consider in this work.
From a GNN perspective, we are interested in learning \textit{asymmetric} relational operators that project the features of each interacting node to a high-dimensional space considering their pairwise neighbors' features to obtain a richer representation at the output layer.
We use a graph \textit{encoder} to compute such representations for each of the meta-type nodes using the HIGs.
The high-dimensional representations are further processed through a \textit{decoder} unit by operating with a set of newly introduced node types to compute the action probabilities and the value function outputs for actor-critic reinforcement learning.
At the decoder level, we introduce an additional low-level interaction graph for this purpose, that we name \textit{heterogeneous decoder graph} (HDG) introducing a set of abstract relations between HIG representations, depots and a value node.

\subsection{The Heterogeneous Mobility Network}
We first introduce three \textit{meta-type} entities for a mobility network, and their corresponding feature spaces.

\subsubsection{Depots}
The mobility network consists of $L$ depots $\mathcal{D} = \{d_1, \dots, d_L \}$ that gets populated with payload requests initiated by clients.
Following the AMoD literature \cite{gammelli2021graph, carron2019scalable}, we define a set of Poisson point processes parameterized by their expected arrival rates $\lambda_d$, $\forall d$ $\in \mathcal{D}$ to model the arrival of payloads at each depot $d$. 
Let $\bar{\alpha}_d \in [\alpha_{min}, \alpha_{max}]$ denote the expected size of a payload requested at depot $d$.
The feature vector of a depot takes form $h_{d}^t = [ x_d, \lambda_d, \bar{\alpha}_d] \in \mathbb{R}^4$, where $x_d \in \mathbb{R}^2$ is the location of the depot.
The features $\bar{\alpha}_d$, $\lambda_d$ can be considered as the vehicle agents' prior knowledge on each depot, that help an agent choose a depot intuitively, even when they are not fully observable, resembling the human taxi driver behavior.

\subsubsection{Payloads}
A payload characterizes a deliverable available at a depot destined to a specific client location.
The set of payloads currently available at a depot $d$ can be denoted as $\mathcal{P}_d^t = \{ p^{dc}_i | i = 1,\dots ,p\_max\}$ where $p\_max \in \mathbb{N}^+$ is a fixed maximum number of payload requests handled by a depot.
Each payload request $p^{dc} \in \mathcal{P}_d^t$ further contain a maximum payoff awarded to the vehicle upon completing the delivery $\mathrm{Payoff}(p^{dc})$, according to the delivery distance and the payload size.
We represent the feature vector of a payload $p^{dc}$, $\in \mathcal{P}_v^t$ by concatenating the payoff, client destination and the required minimum vehicle capacity. 
Thus, $h_p = [\mathrm{Payoff}(p^{dc}), x_c, \mathrm{Cap}(p^{dc})] \in \mathbb{R}^4$.
The incoming payload requests are inserted to the corresponding depot's payload queue $\mathcal{P}_d$ ordered by their arrival time.
The payload assignment function $\Psi$ follow a fixed policy that return the next suitable payload for the requesting vehicle from the payload queue, to help minimizing the waiting time of a payload request.

\subsubsection{Vehicles}
We define the feature space of a vehicle $v$ as the vector $h_{v}^t = [x_{v}^{t_{prev}}, x_{v}^{t_{next}}, \mathrm{Cap}(v)] \in \mathbb{R}^5$, where $x_{v}^{t_{prev}}$,$x_{v}^{t_{next}} \in \mathcal{D} \cup \mathcal{C}$ are previous and the next stops of the vehicle. 

The observation feature space of a vehicle concatenates the meta-type features, thus $\mathbf{h}_v^t = [h_{d}^t, h_{p}^t, h_{v}^t]$. 

\subsection{Time-Varying Heterogeneous Interaction Graph}
\begin{figure}[t]
    \centering
    \includegraphics[trim={0.1cm 0.1cm 0.1cm 0.1cm}, clip, width=0.25\textwidth]{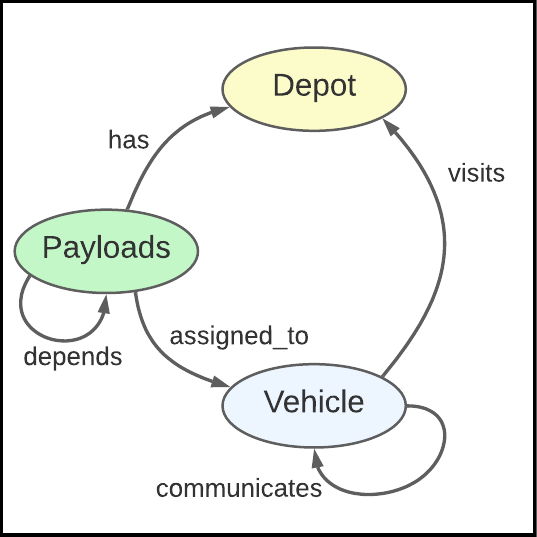}
    \caption{The meta-graph representing the abstract interactions among vehicle, depot and payload type entities. The vehicle and payload types has self-edges that connects the objects of these types to themselves. }
    \label{fig:hig_meta}
    \vspace{-10pt}
\end{figure}
We introduce five semantic relations to summarize the interactions among different meta-type objects: $\Phi = \{\mathrm{has}$, $\mathrm{visits}$, $\mathrm{depends}$, $\mathrm{assigned\_to}$, $\mathrm{communicates}\}$.
The observable neighborhood of a vehicle $\mathcal{N}_{v}^t = {\{ \mathcal{V}_v^t, \mathcal{D}_v^t \}}$ follows a topological range, where $\mathcal{V}_v^t =$ $\{v'| \forall v' \in \mathcal{V}$,$\mathrm{Distance}(v, v') \leq \mathrm{Distance}(v, v_{k_v})\}$,
and $\mathcal{D}_v^t = \{d| \forall d $ $\mathrm{Distance}(v, d) \leq \mathrm{Distance}(v, d_{k_d}) \}$.
Here $v_{k_v}$, $d_{k_d}$ denotes the $k_{.}-$th closest vehicle and the depot respectively.
Let $\mathcal{P}_v^t$ define the set of vehicle $v$'s all observable payloads where $\mathcal{P}_v^t = \{ \mathcal{P}_d^t | \forall d \in \mathcal{D}_v^t \}$. 
We construct the HIG for a vehicle $v$, $\mathcal{G}_v^t$ for timestep $t$, by including the observable vehicles, payloads and  depots. 

Algorithm \ref{algo:hig} summarizes the steps for connecting the each type node using semantic relations following the meta-graph shown in Fig. \ref{fig:hig_meta}.
The $\mathrm{communicates}$ edge captures interactions between any two vehicles in the neighborhood that includes itself $\mathcal{V}_v^t$, allowing the vehicle type nodes to incorporate each others features.
Next, each vehicle node relates to the depots through a $\mathrm{visits}$ type edge, which operates the vehicle agents' features with the depot features --both observed and prior knowledge, to compute the output encoding.
The payload nodes in HIG connects to their corresponding depots through a $\mathrm{has}$ type edge. 
By taking into account each observable payload's minimum-required capacity, we add an $\mathrm{assigned\_to}$ type edge between any the matching vehicles and the payloads, through line 16 to 21 in Algorithm \ref{algo:hig}.

The intuition behind drawing semantic relations not only for the \textit{ego} vehicle node $v$, but also for neighboring vehicles in HIG narrows down to two objectives: 1) $v$'s objective to fictitiously approximating the other agents' actions, and 2) learning robust relational operators with limited information; especially, by accounting for the variance when the neighbors' higher-order relations are not observable.
Note that any meta-type node that does not have an incoming edge is not passed through the convolution layers in graph neural networks.
Thus, we add a self-edge connection $\mathrm{depends}$ to operate the payloads' type node feature space with themselves.
These incoming edges allow aggregating the features from neighboring objects, resulting in richer node representations at the output layer.

\begin{algorithm}
\SetAlgoLined
 \textbf{Inputs:} $\mathcal{N}_{v}^t =\{\mathcal{V}_v^t,\mathcal{D}_v^t\}$, $\mathcal{P}_v^t$, $\mathcal{D}$\\
 \textbf{Output:} $\mathcal{G}_v^t$\\
\For{$v \in \mathcal{V}_v^t$}{
    \For{$d \in \mathcal{D}$}{
            Add Edge ($v$, $\mathrm{visits}$, $d$)
    }

    \For{$v' \in \mathcal{V}_v^t$}{
        Add Edge ($v$, $\mathrm{communicates}$, $v'$)
    }
}


\For{$d \in \mathcal{D}_v^t$}{
    \For {$p^{dc} \in \mathcal{P}_d^t$}{
        Add Edge ($p^{dc}$, $\mathrm{has}$, $d$)
    }
}
\For {$p^{dc} \in \mathcal{P}_v^t$}{
    \For{$v \in \mathcal{V}_v^t$}{
        \If{\big [ $\mathrm{Cap}(p^{dc}) 
        \leq \mathrm{Cap}(v)$  \big ]}
        {
        Add Edge ($p^{dc}$, $\mathrm{assigned\_to}$, $v$)
        }
    }
    
    Add Edge ($p^{dc}$, $\mathrm{depends}$, $p^{dc}$)
}


Create graph $\mathcal{G}_v^t$ with edges.
 \caption{Constructing the HIG: $\mathcal{G}_v^t$}
 \label{algo:hig}
\end{algorithm}

\subsection{Representation Learning with HetGAT}

A HetGAT layer intakes one's initial features $h_i$, according to the HIG to compute a high-dimensional projection $h'_i$ by applying node-wise \textit{message passing}, \textit{aggregation} and \textit{attention} operations at each node.
Since we operate on the HIG and a feature set observed by a vehicle at a given timestep, we drop the timestep $t$ for brevity. 
In the GNN message passing, each node propagates its feature vector to the neighboring nodes $\mathscr{N}_i$ following the directionality ascribed in the relation preserving the asymmetry. 
Note that $\mathscr{N}_i$ is the first order neighborhood of some meta-type node $i$ in the HIG in $\mathcal{G}$, which is different from the observational neighborhood of a vehicle $\mathcal{N}_{v}$.
The features are then multiplied with relation-specific weight matrices to project them to a high-dimensional feature space.
As the weight matrices are relation specific, they are generalizable to different input graph sizes, in contrast to fully-connected neural networks that depend on the input size.


Let $\mathrm{Type}(i,j)$ denote the type of edge between $i,j$ where $\mathrm{Type}(i,j) = \phi \in \Phi$, and $i,j \in \mathscr{N}_i$.
For projecting the feature spaces of different sizes to the output shape $h'_i$, the weight matrices are shared in a relation-specific manner. 
For any $\mathrm{Type}(i,j) = \phi$, where $j \in \mathscr{N}_i$, we define $W_\phi$'s dimensions as $|h'_i|\times|h_j|$, where $W_\phi$ is a projection weight matrix shared among the nodes participating in relation $\phi$.
The node-wise message passing in a single HetGAT layer can be summarized as,
\begin{equation}
    \bar{h}^\phi_i = \sigma \Big [\sum_{\substack{j \in \mathscr{N}_i \\ \mathrm{Type}(i,j) = \phi}} \beta_{ij} W_\phi h_j\Big],
\end{equation}
where $\beta_{ij}$ is a node-wise attention coefficient, and $\sigma$ is a non-linear activation function.
A node $i$ may have incoming messages over different edges; i.e., a depot type node receives messages over $\mathrm{has}$, and $\mathrm{visits}$ type edges.
In such cases, we aggregate each feature message using a rotational invariant operation $\mathrm{Agg}$.
Thus, we denote the outgoing feature space $h'_i$ as
\begin{equation}
    h'_i = \mathrm{Agg}\Big(\bar{h}^{\phi_1}_i \dots \bar{h}^{\phi_n}_i \Big),
    \label{eq:agg}
\end{equation}
where $n$ is the number of distinct incoming edge types for node $i$.
In this work, we use Leaky ReLU activation for $\sigma$ and mean aggregation for $\mathrm{Agg}$.
The node-wise attention weights $\beta_{ij}$ emphasize the importance of the neighbor $j$'s features to $i$'s action selection.
Briefly, HetGAT learns an attention \textit{coefficient} $e_{ij}$ via a fully-connected layer $\mathrm{fc}$ parameterized by an edge-specific weight matrix, and LeakyReLU activation, $\mathrm{fc}: \mathbb{R} ^{2|h'_i|}$$\mapsto$$\mathbb{R}$.
Thus, for a given relational edge type $\phi$
\begin{equation}
    e_{ij} = \mathrm{fc} \big ( W_\phi h_i, W_\phi h_j \big. ).
\end{equation}
Finally, the attention coefficients are normalized over the neighborhood $\mathscr{N}_i$ using the softmax function as
\begin{equation}
    \beta_{ij} = \frac{\exp(e_{ij})}{\sum_{k \in \mathscr{N}_i} \exp(e_{ik})}.
\end{equation}

We stack multiple HetGAT layers to learn high-dimensional representations for each node in the input HIG.
Thanks to the high expressiveness and the ability to represent diverse array of entities engaging in complex agent interactions, we believe that HetGAT-based approaches are ideal for learning many heterogeneous fleet coordination tasks. 


\subsection{Graph Attention Policy Architecture}
\begin{figure}[t]
    \centering
    \includegraphics[trim={0cm 0cm 0cm 0cm}, clip, width=0.48\textwidth]{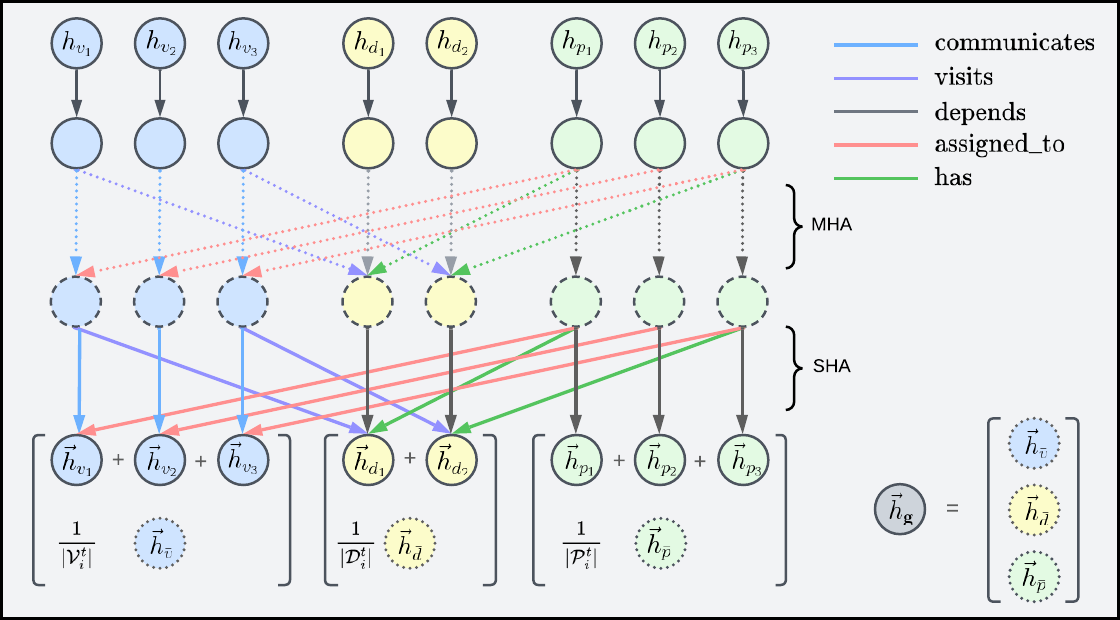}
\caption{The HetGAT encoder architecture. Following Fig. \ref{fig:hig_meta} the blue, yellow and green colors represent vehicle, depot and payload meta-type objects in the input HIG. Each relation type is represented in corresponding colors. The HIG is first sent through multiple multi-head attention (MHA) layers and finally a single-head attention (SHA) layer. The graph node embedding is represented in grey color by stacking the \textit{mean nodes} of each meta-type outputs embeddings.}
\label{fig:encoder}
\vspace{-10pt}
\end{figure}
We consider two criteria to assess the generalizability of a trained policy in mobility environments: 1) transferability to different mobility networks to the one that it was trained on with minimal to no reconfiguration, and 2) shareability by different vehicle classes, i.e., capacity, to maximize the rewards.
Primarily, a HetGAT's generalizability attributes to its type-specific sharing of graph convolution and attention operators which makes them operable on arbitrary graph sizes with nodes of different degrees \cite{kipf2016semi}. 
In heterogeneous mobility networks, such interaction graphs often occur as a result of 1) time-varying observability and 2) the addition and removal of different meta-type nodes to cater to dynamic demand patterns.
\rebutt{
In \cite{kool2018attention}, authors showed generating attention-based graph encoding at the node level is highly effective in solving routing problems.
Following this notion, we propose a novel type-sensitive HetGAT encoder-decoder architecture for solving autonomous mobility.
Generally in deep learning, the encoder unit processes the input to construct rich representations from the inputs, whereas the decoder generates the overall output conditioned on the encoder output \cite{goodfellow2016deep}. 
Similarly in graph learning, an encoder may generate high-dimensional graph representations at the output layer, where the decoder uses encoder outputs to perform prediction or classification \cite{kazemi2020representation, zhu2022learnable}.
}

\subsubsection{Encoder}
The encoder unit in this work intakes the HIG $\mathcal{G}_v$ and the associated features of the nodes.
The graph is then passed through 2 multi-head attention (MHA) and a single head attention (SHA) output layers.
As presented in \cite{velivckovic2017graph}, a MHA layer computes $k_\beta$ independent attention weights and concatenates the aggregated features in the outgoing feature space $h'_i$ resulting an output dimensionality $k_{\beta}|h'_i|$ compared to the single-head attention (SHA) presented in Eq. \ref{eq:agg}.
Fig. \ref{fig:encoder} shows the proposed encoder architecture.
For each meta-type node in the output representation $\Vec{h}_{d}$, $\Vec{h}_{v}$ and $\Vec{h}_{p}$ we use $\mathbb{R}^{64}$ vectors.
In addition to each meta-type node representations, the encoder outputs a \textit{graph embedding} node shown in grey color $\mathbf{g}$ by averaging each meta-type node and concatenating them together, where $h_{\mathbf{g}} \in \mathbb{R}^{|\Vec{h}_v|+|\Vec{h}_d|+|\Vec{h}_p|}$.
\begin{algorithm}
\SetAlgoLined
 \textbf{Inputs:} $\mathbf{g}$, $\mathcal{D}$, $\mathbf{val}$\\
 \textbf{Output:} $\mathcal{G}_{dec}$\\
Add Edge ($\mathbf{g}$, $\mathrm{g\_contributes\_val}$, $\mathbf{val}$) \\
\For {$d \in \mathcal{D}$} {
        Add Edge ($d$, $\mathrm{d\_contributes\_g}$, $\mathbf{g}$) \\
        Add Edge ($d$, $\mathrm{d\_contributes\_val}$, $\mathbf{val}$) \\
        \For {$d' \in \mathcal{D} $} {
           Add Edge ($d$, $\mathrm{d\_near\_d}$, $d'$) \\
        }
    }
Create graph $\mathcal{G}_{dec}$ with edges.
 \caption{Constructing the HDG: $\mathcal{G}_{dec}$}
 \label{algo:hdg}
\end{algorithm}

\subsubsection{Decoder}
Let $\mathbf{g}$, $\mathbf{val}$ denote the graph embedding node and a newly introduced value node.
We summarize the steps of constructing the heterogeneous decoder graph (HDG) in Algorithm \ref{algo:hdg}.
The decoder accepts the HDG along with the graph embedding node $\Vec{h}_{\mathbf{g}}$, depot representations $\Vec{h}_{d}$, and $\mathbf{val}$ node --a zero vector for value node initialization.
The decoder processes the HDG through two HetGAT layers where the first layer has MHA and an output layer with SHA.
We provide the details of chosen output feature dimensions of each HetGAT layer in Appendix \ref{app:nn}.
The value node output from the decoder $\Vec{q}_{val}$ is further processed through a fully connected layer $\mathrm{fc\_{val}}$ to obtain the value function output $q_{val}$.
At the final layer, we do not perform feature aggregation and non-linear activation steps for output graph embedding and depot nodes.
Instead, each depot node output is dot multiplied with the graph embedding node output to compute the output query values followed by a non-linearity; $q_d = \sigma (\Vec{q}_{d}^T\Vec{q}_{\mathbf{g}})$ for all $d \in \mathcal{D}$, where $q_d$ is the value of choosing depot $d$.
Finally, we calculate probabilities associated with each depot in the stochastic policy by using the softmax function over each $q_d$.

\subsubsection{Fleet Rebalancing Mask} 
In the absence of suitable payloads nearby, one must favor farther away depots to avoid getting penalized by choosing an empty depot.
Following this notion, we draw parallels between a non-repopulating mobility environment and a \textit{stochastic variant} of reward-collecting travelling salesman problem (RC-TSP).
In contrast to the RC-TSP, where salesman's overall reward depends on a set of node-specific values that decays upon visitation, an stochastic variant may suffer from the added difficulty of partial observability and multiple salesmen moving simultaneously, resembling the mobility game.
In \cite{kool2018attention}, authors show that masking is beneficial in solving RC-TSP to prevent visiting an already visited node and getting penalized.
Thus, we introduce a fleet rebalancing mask computed using local observations to 1) explore farther away depots in low-demand environments, and 2) prevent one from choosing depots in the observable range that does not contain matching payloads.

Formally, we mask the query values of each depot that is in the observation range, but does not contain a suitable payload such that, $q_d = -\infty$ for all $d \not\in $ $\{d | \forall p \in \mathcal{P}_d, \mathrm{Cap}(p) \leq \mathrm{Cap}(v), \forall d\}$.
From a mobility perspective, we resemble this to an intrinsic \textit{fleet rebalancing} mechanism.


\begin{figure}[t]
    \centering
    \includegraphics[trim={0cm 0cm 0cm 0cm}, clip, width=0.4\textwidth]{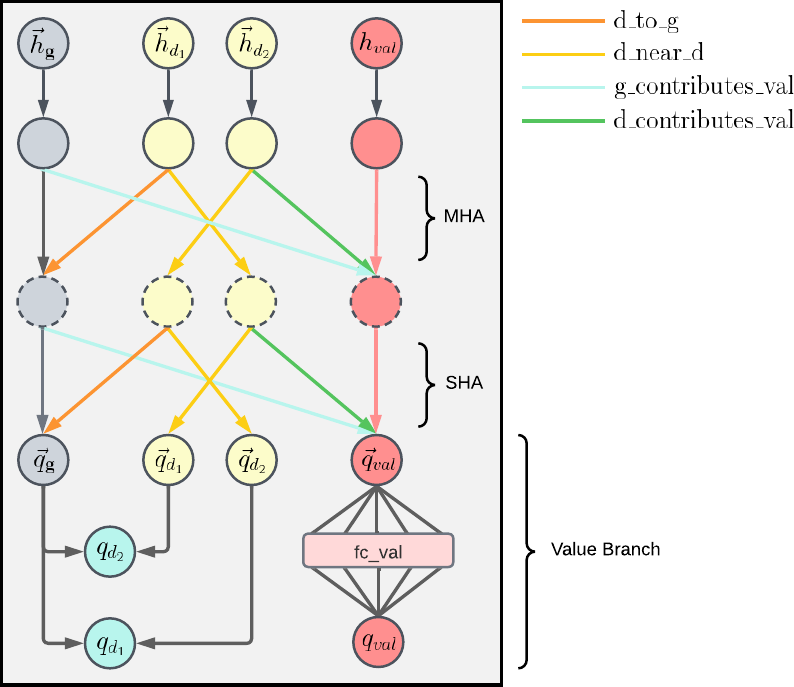}
\caption{The HetGAT decoder architecture. The critic value function shares layers with the actor network, yet the value branch is only used by the critic network. The final graph and the depot embeddings are multiplied together to output the action-values of choosing a depot $q_{d_i}$. }
\label{fig:decoder}
\end{figure}


\section{Experiments and Results}
\label{sec:experiments}

\begin{figure*}[t]
    \centering
    \subfigure[]
    {\includegraphics[trim={0cm 0cm 0cm 0cm}, clip, width=0.23\textwidth]{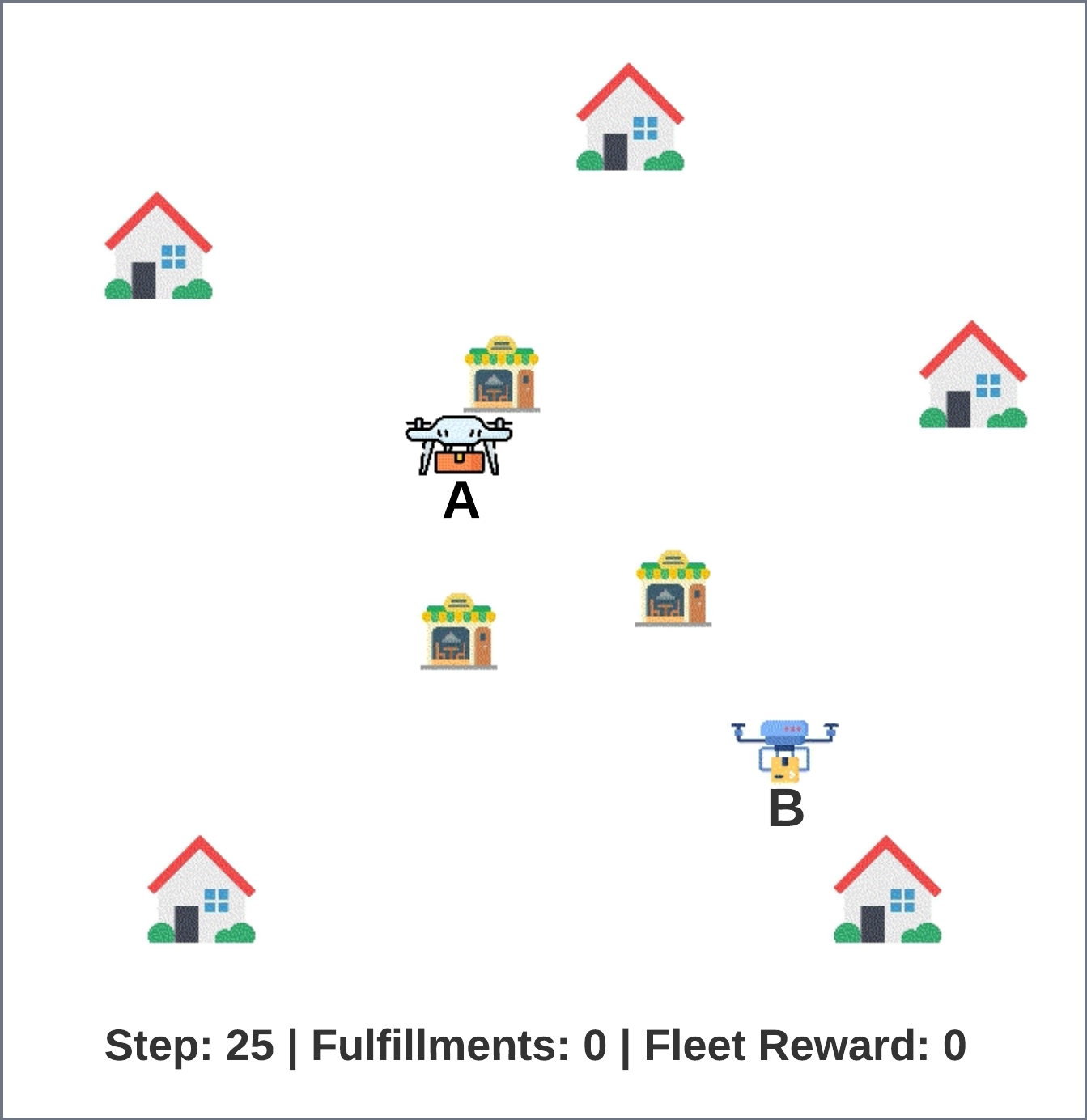}
    \label{fig:sim1}}
    \subfigure[]
    {\includegraphics[trim={0cm 0cm 0cm 0cm}, clip, width=0.23\textwidth]{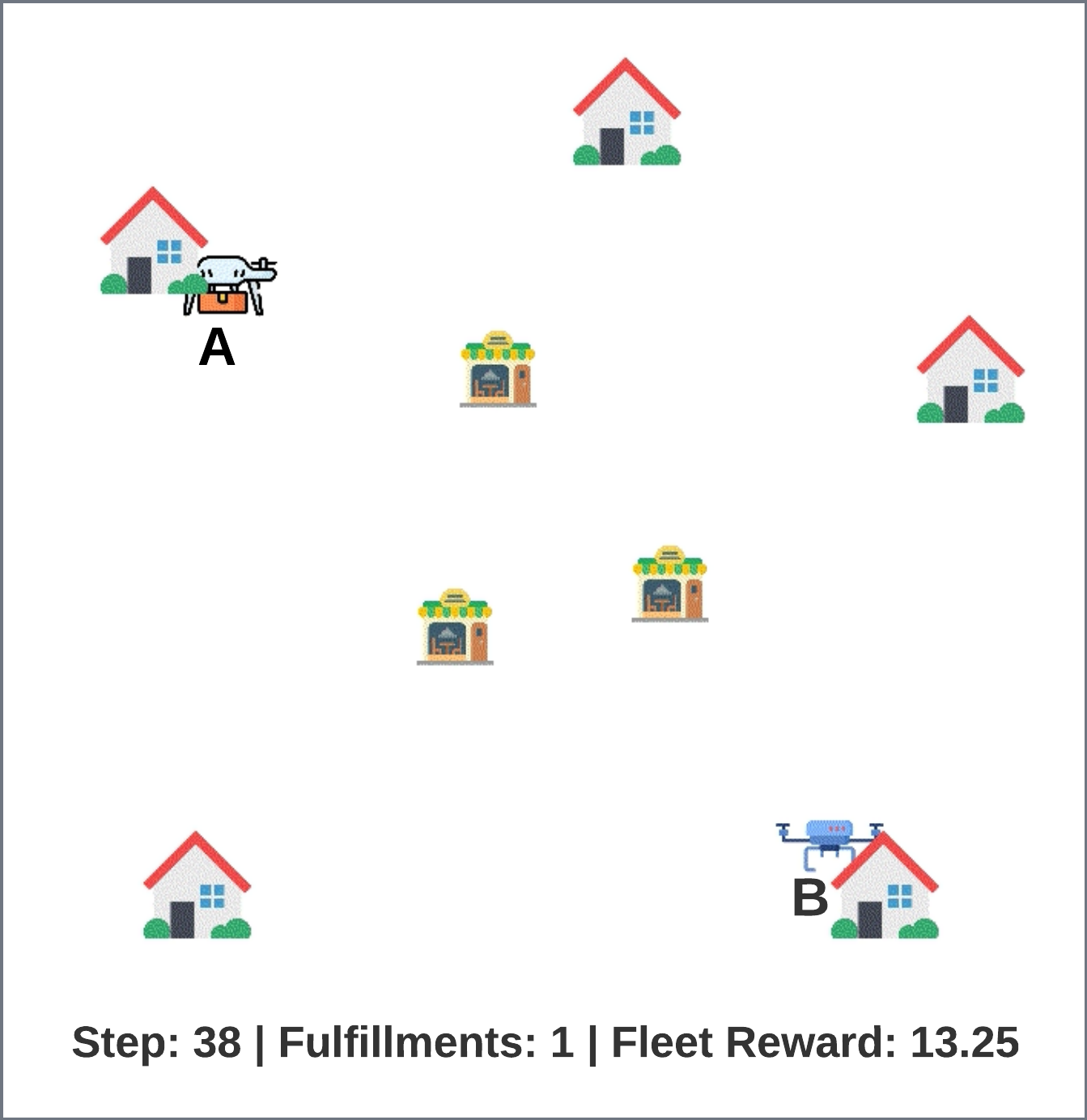}
    \label{fig:sim2}}
    \subfigure[]
    {\includegraphics[trim={0cm 0cm 0cm 0cm}, clip, width=0.23\textwidth]{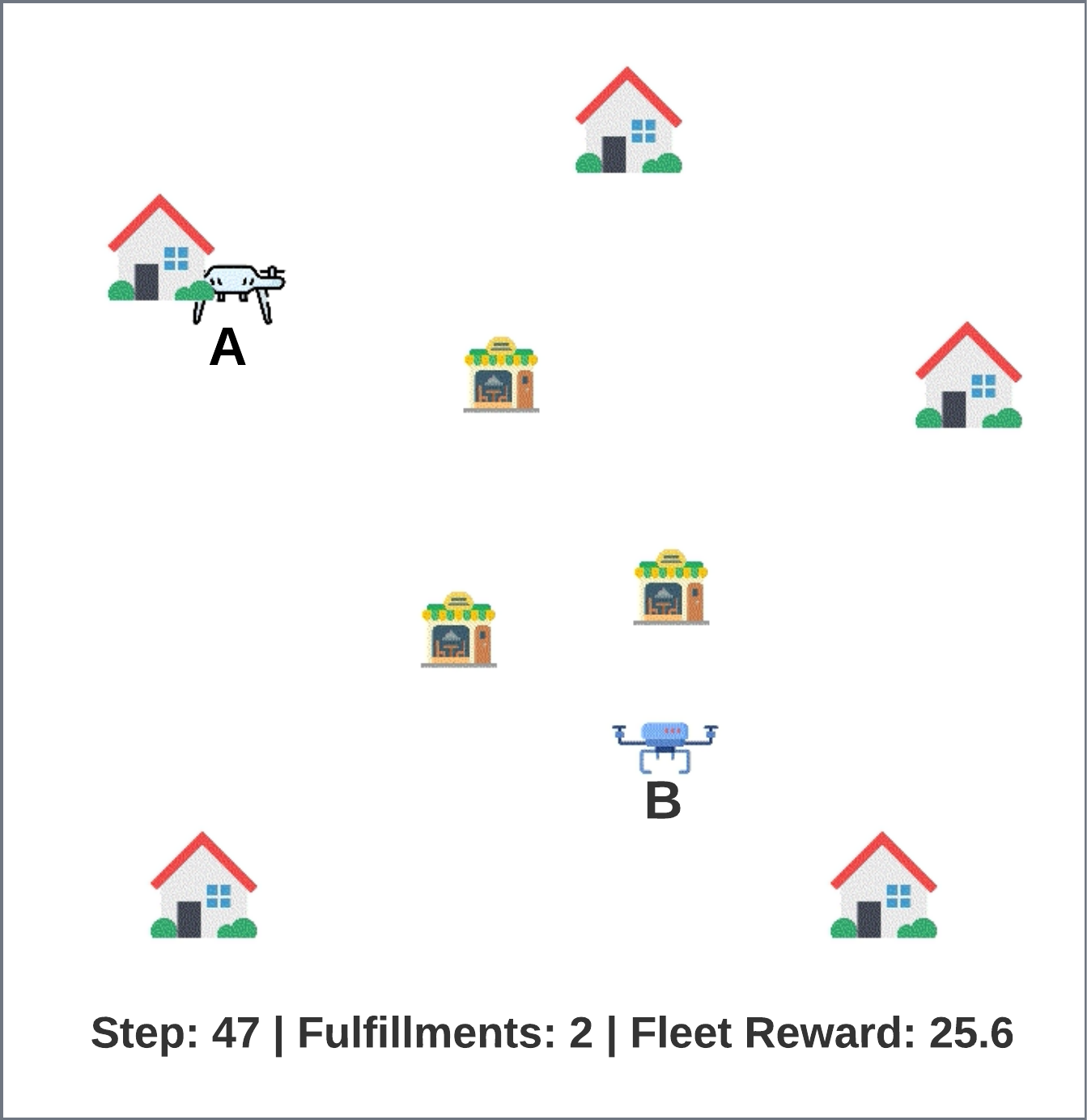}
    \label{fig:sim3}}
    \subfigure[]
    {\includegraphics[trim={0cm 0cm 0cm 0cm}, clip, width=0.23\textwidth]{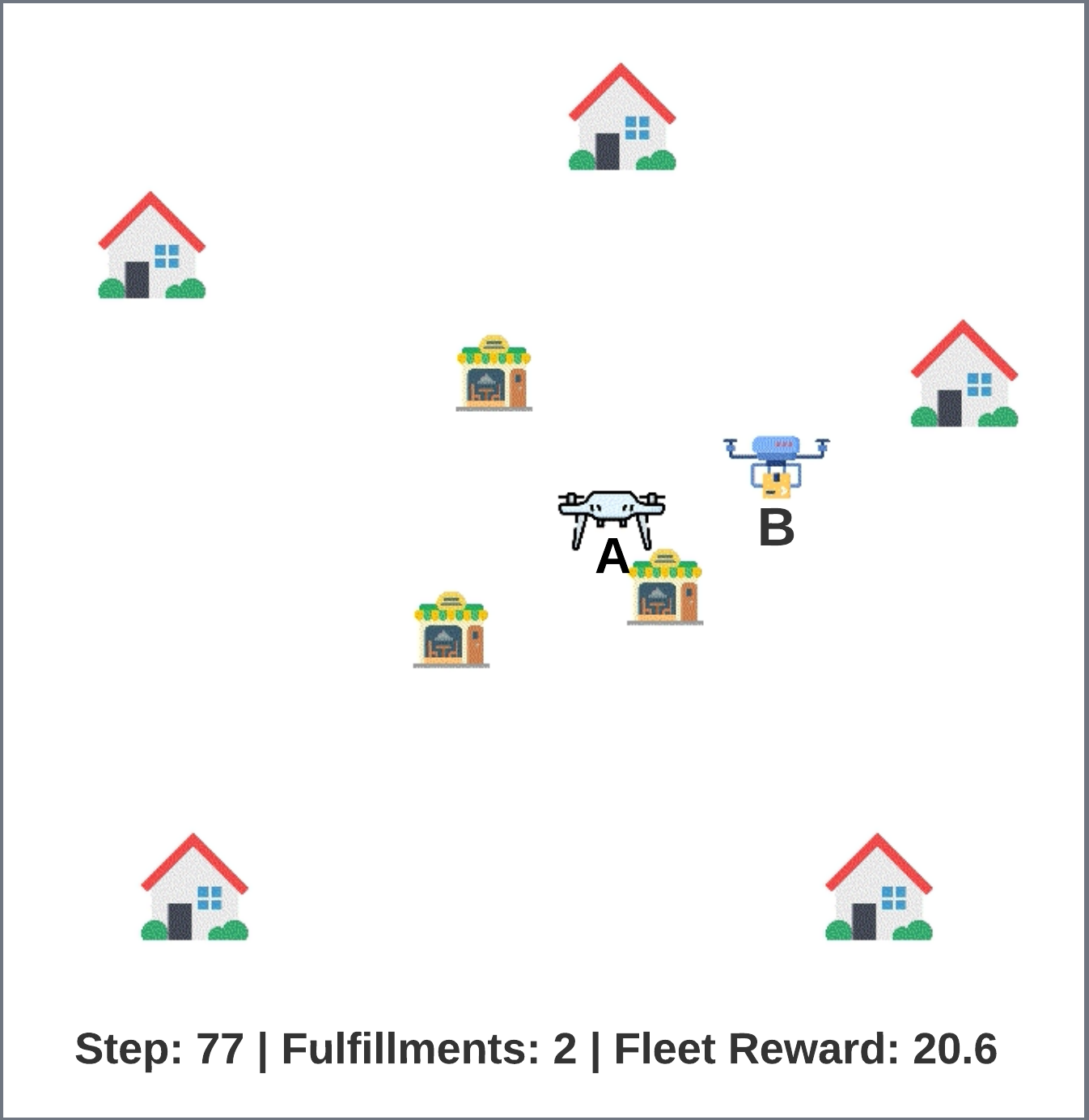}
    \label{fig:sim4}}
    
    \caption{Snapshots of a simple AAM simulation with 2 different vehicles, 5 clients, and 3 depots. \textbf{(a)} Both the vehicles are delivering payloads picked up at their initial depots to the clients. \textbf{(b)} An active timestep: \textit{B} reaches the destination client, drops off the payload and collects a positive reward. \textit{A} is still completing the delivery. \textbf{(c)} The next active timestep: \textit{A} drops off the payload and collects a positive reward. \textit{B} travels to a chosen depot to pick up another payload. \textbf{(d)} The next active timestep: \textit{A} reaches an invalid depot, and collects a reward of -5. \textit{B} is on its way to deliver another payload. Elapsed $\Delta t$ timesteps, total fulfillments, and the total fleet reward are displayed at the bottom of each image.}
    \label{fig:sims}
\end{figure*}

\begin{figure}[t]
    {\includegraphics[trim={0cm 0cm 0cm 0cm}, clip, width=0.48\textwidth]{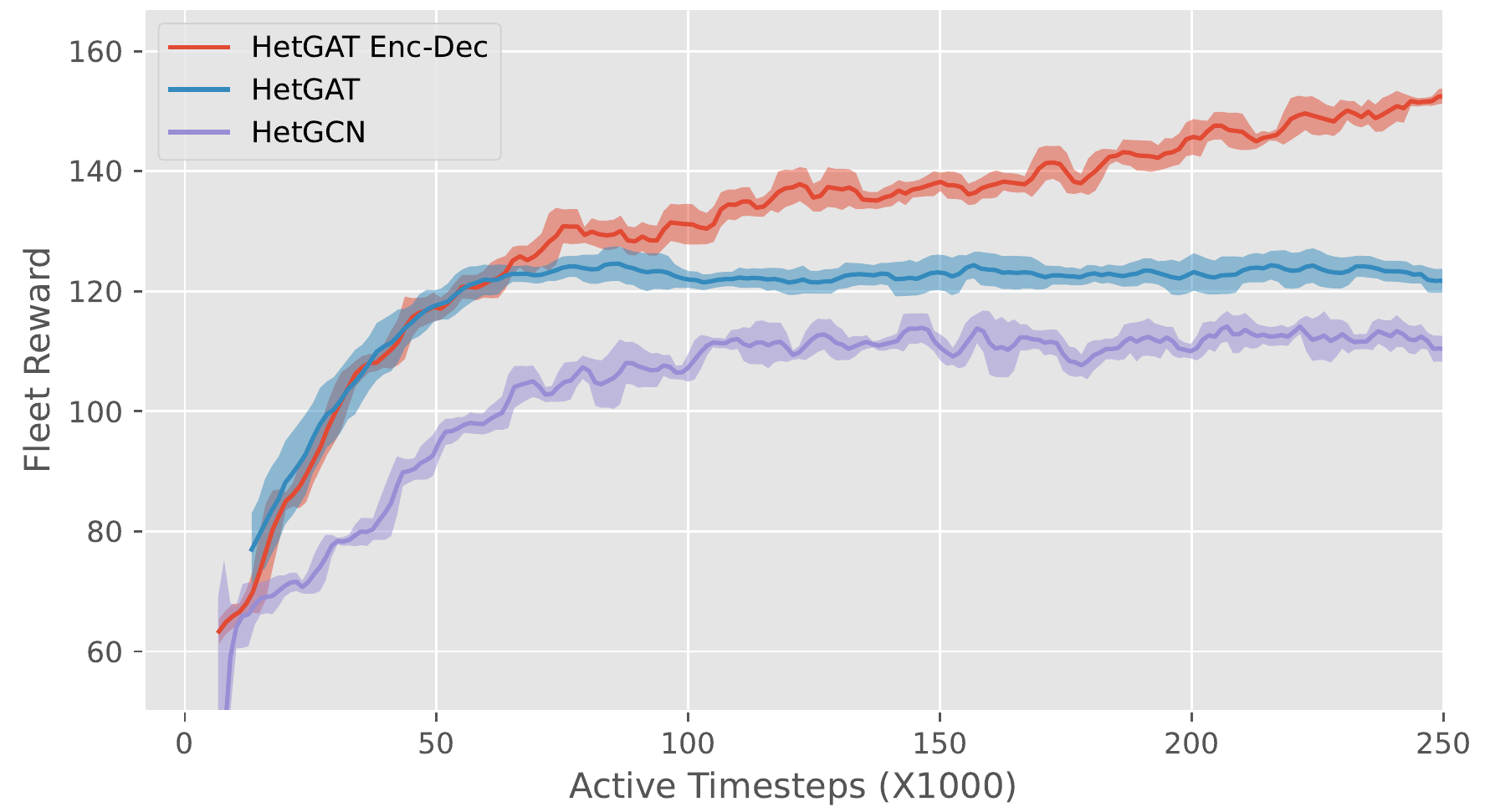}}
    \caption{Training performances of HetGAT Enc-Dec architecture compared to HetGAT and HetGCN after 250000 active timesteps. A heterogeneous fleet of 6 vehicles, 2 vehicles from each category, 10 depots and 12 client nodes were used for the training.}
    \label{fig:training}
\end{figure}

\subsection{Simulation Environment}
We implemented the AAM environment on PettingZoo framework \cite{terry2021pettingzoo}, HetGAT Enc-Dec using the Deep Graph Library (DGL) with a PyTorch back-end, and performed the MARL using Ray RLlib \cite{liang2017ray} to scaling the training process.
We trained our system on an NVIDIA A100 GPU and an AMD EPYC 7713 processor for 10 hours on Indiana University \textit{Big Red 200} computing facility.
\footnote{It is also possible to train the HetGAT Enc-Dec policy on a desktop computer with an NVIDIA RTX 3090 GPU and an Intel 12700K CPU under 10 hours.}

For evaluating the proposed approach, we consider a custom AAM environment where a payload destination can either be a client or a depot node; thus a modified client set is $\mathcal{C}'$ = $\mathcal{D} \cup \mathcal{C} 
\setminus d$ and $c \in \mathcal{C}'$.
We categorize the payloads and vehicles into three different sizes such that $\mathrm{Cap}(v)$, $\mathrm{Cap}(p^{dc})$, $\in \{1,2,3\}$,
where $\mathrm{Cap}(v) = 3$ denotes the largest of the UAVs that can carry any payload, and $\mathrm{Cap}(v) = 1$ indicates the smallest that can only carry payloads of size 1.
In the experiments, we do not consider the scenario of a UAV carrying multiple payloads at once.
Throughout the experiments, we used $p\_max=5$ as the maximum payload queue length of a depot.
The vehicles use a constant velocity trajectory to navigate to their destinations.
For the on-demand scenario we consider a simulation episode of 400 $\Delta t$ timesteps, and a horizon length $\mathbf{T} = 50$ active timesteps where the agents are allowed to take actions.
During the training, we skip non-active timesteps to improve efficiency and prevent the training algorithm from accumulating unnecessary observations.
The training environment comprise of 24 $\times$ 24 discrete cells where the depot and client nodes are approximately evenly positioned in each quadrant.
Fig. \ref{fig:sims} shows a simulation environment with 2 vehicle agents delivering payloads to clients, where Fig. \ref{fig:ss_full} shows a densely populated simulation environment.

\subsubsection{Populating Payloads}
The payload requests may arrive at depot $d$ every 50 $\Delta t$ intervals following a Poisson process whose rate parameter is chosen uniformly from the set $\{0.01, 0.05, 0.025\}$.
We have chosen the rate parameters to reflect the imbalanced request arrival nature at different depots in a city, where a higher parameter may simulate the behavior of a high-demand depot.
For every incoming request $p^{d.}$ we assign a capacity $\mathrm{Cap}(p^{d.})$ by sampling from the normal distribution $\mathrm{Normal}(\bar{\alpha}_d,0.1)$, such that the rounded $\mathrm{Cap}(p^{d.}) \in \{1,2,3\}$. 
Here $\bar{\alpha}_d$ is the expected payload size at depot $d$.
Typically, in a practical last-mile delivery system, it is more likely that the payload requests arriving at a depot require delivering to a destination closer to the origin than those further away for various reasons, including minimizing the carbon footprint.
We mimic this behavior in the training by ordering the nodes by their distance, in a normally distributed manner.

\subsubsection{Reward Function}
\label{sec:reward}
The maximum payoff that an agent can obtain by delivering a payload $\mathrm{Payoff}(p^{dc})$ depends on the distance between the origin and the destination, and the payload size.
Specifically, we define the payoff as a nonlinear function of the distance using the taxi fare computation scheme proposed in \cite{yang2010nonlinear}.
In practice, this prevents the vehicles from unfairly gaining high payoffs that linearly increase with the distance. 
We observed that non-linear rewarding scheme to 1) discourage certain depots with longer rides from emerging as dominating actions in the game from preventing asymmetric vehicle distribution, 2) stabilize the training process.
Thus,
\begin{equation}
 \mathrm{Payoff}(p^{dc}) = \mathbf{q_1}||x_d-x_c||^2 + \mathbf{q_2}||x_d-x_c|| + \mathbf{q_3}\mathrm{Cap}(p^{dc}), 
 \end{equation}
for $d \in \mathcal{D}$, $c \in \mathcal{C}'$, $\mathbf{q_1} < 0$, and $\mathbf{q_2}, \mathbf{q_3} > 0$.
Specifically, under this payoff scheme, the vehicles are incentivized in a concave fashion rather a linear fashion, thus selecting farther depots is not always preferred.
The payload size acts as a \textit{flag fall cost} multiplied by $\mathbf{q_3}$ in the payoff function.
The \textit{net reward} $r_v$ of $v$ for choosing a depot is the difference between the maximum payoff and the vehicle's travel cost to reach the depot for picking up the payload (Eq. \ref{eq:rew}). Thus,
 \begin{equation}
 r_v^t = 
  \begin{cases}
  \mathrm{Payoff}(p^{dc}) - \mathbf{q_4}||x_{v}^t - x_{d}||, \hspace{20pt} \text{if } d \text{ is valid,} \\
   \hspace{25pt} 0 \hspace{48pt} \text{if } d \text{ is invalid and } x_{d} = x_{v}, \\
    \hspace{18pt} -5 \hspace{123pt} \text{otherwise.}
\end{cases}
 \label{eq:rew}
\end{equation}

In other words, if the vehicle chooses a depot that returned a suitable payload, it may complete the delivery and obtain a reward according to the first case.
In the cases where the chosen depot does return with a suitable payload, the vehicle may receive a penalty of $-5$ rewards, except when the chosen depot is its current location. 
By considering the distance between the nodes in the mobility network, we chose the coefficients $\mathbf{q_1} = -0.0167$, $\mathbf{q_2} = 1$, $\mathbf{q_3} = 2$, and $\mathbf{q_4} = 0.2$ to flatten the concave payoff curve at a maximum trip distance of 30 units. 
For more information on calculating the coefficients we refer the readers to \cite{yang2010nonlinear}.

\subsection{One-Shot Training and Comparison}
We observed training the agents directly in the on-demand mobility environment to cause a skewed behavior where agents largely preferred busier depots, causing an imbalance in the mobility network.
Inspired by the RC-TSP, we propose a step-wise approach: first training the vehicle agents in an environment that only gets populated once, --that we call ``one-shot training", and simulating until either all the payloads are delivered or reach a fixed duration of 100 $\Delta t$ timesteps.
This encourages the agents to continuously fulfilling the payloads in an increasingly scarce environment, requiring them to take more exploratory actions to minimize penalization.

We used a fixed-size heterogeneous vehicle fleet that comprises 6 vehicles -- 2 from each capacity, and a fixed observation range of $k_v$ = 5 and $k_d$ = 5, 10 depots and 12 clients for one-shot training (highlighted in in Table \ref{tab:results}).
The expected arrival rates, and the expected payload sizes were randomized to prevent the model from overfitting and to generalize better to different environments.

We compare the performances of HetGAT Enc-Dec to two other GNN architectures: vanilla HetGAT and Heterogeneous Graph Convolutional networks (HetGCN). 
Fig. \ref{fig:training} shows the  total fleet reward, the sum of all the agents' rewards acquired under each policy architecture against the number of active timesteps trained.
During the one-shot training, the HetGAT Enc-Dec policy significantly outperformed the HetGAT and HetGCN MARL policies.
The vanilla HetGAT and the HetGCN module architectures resembled that of the encoder with similar semantic relations showed in Fig. \ref{fig:encoder}, yet the final layer depot representations $\Vec{h}_{d}$ directly corresponded to the action-value output of a depot.
Appendix \ref{app:nn} provides the implementation information for the three neural network architectures and the training hyperparameters. 
Despite being limited in its ability to generalize to different mobility networks, we also experimented Long-Short Term Memory (LSTM) policy network yet, its training performances were significantly worse compared to the others; thus, we exclude it in further evaluations.
Each experiment comprise 20 simulation episodes that lasted 400 $\Delta t$ timesteps.
\begin{figure}[t]
    \centering
    \subfigure[]
    {\includegraphics[trim={0cm 0cm 0cm 0cm}, clip, width=0.23\textwidth]{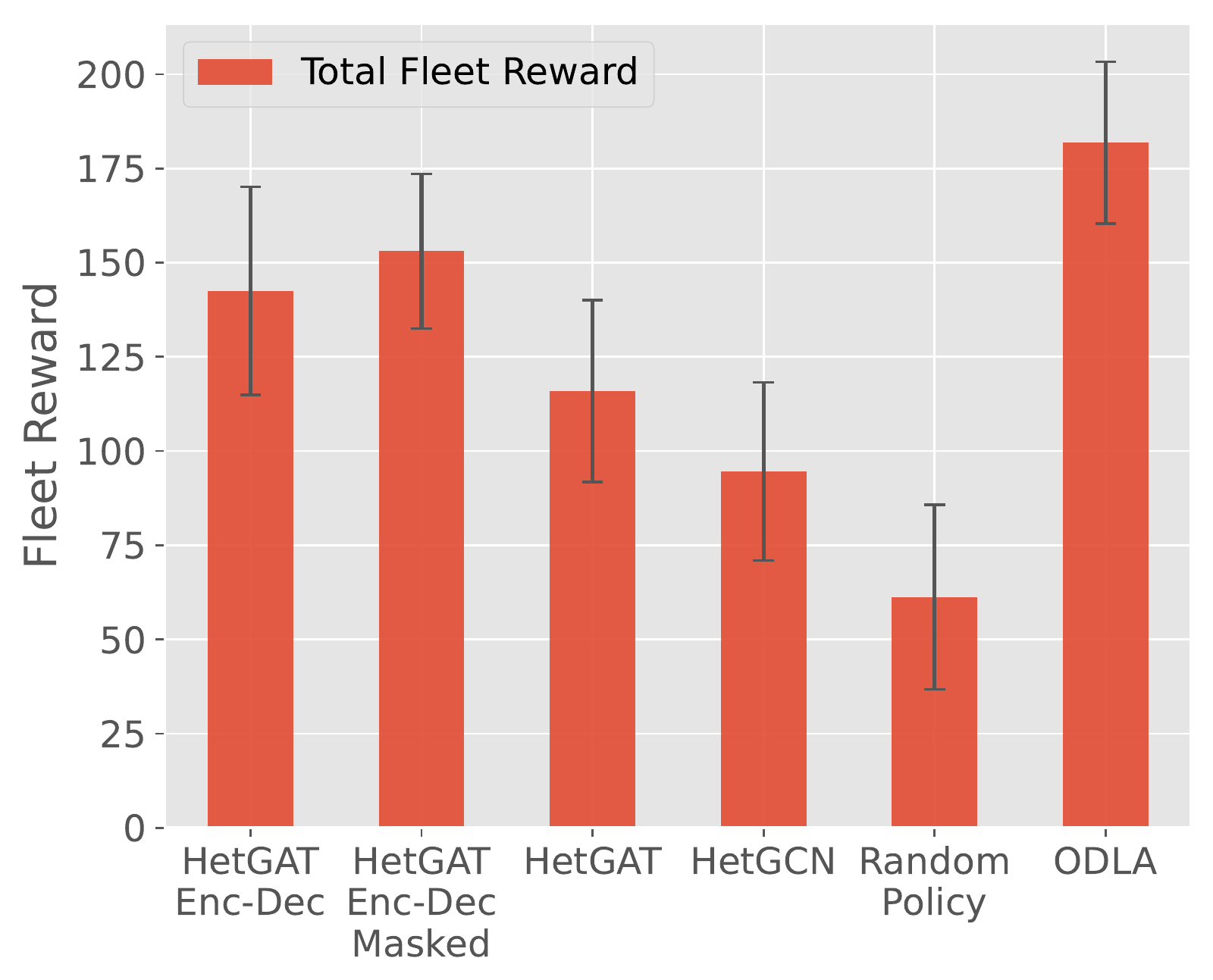}
    \label{fig:eval_com}}
    \subfigure[]
    {\includegraphics[trim={0cm 0cm 0cm 0cm}, clip, width=0.23\textwidth]{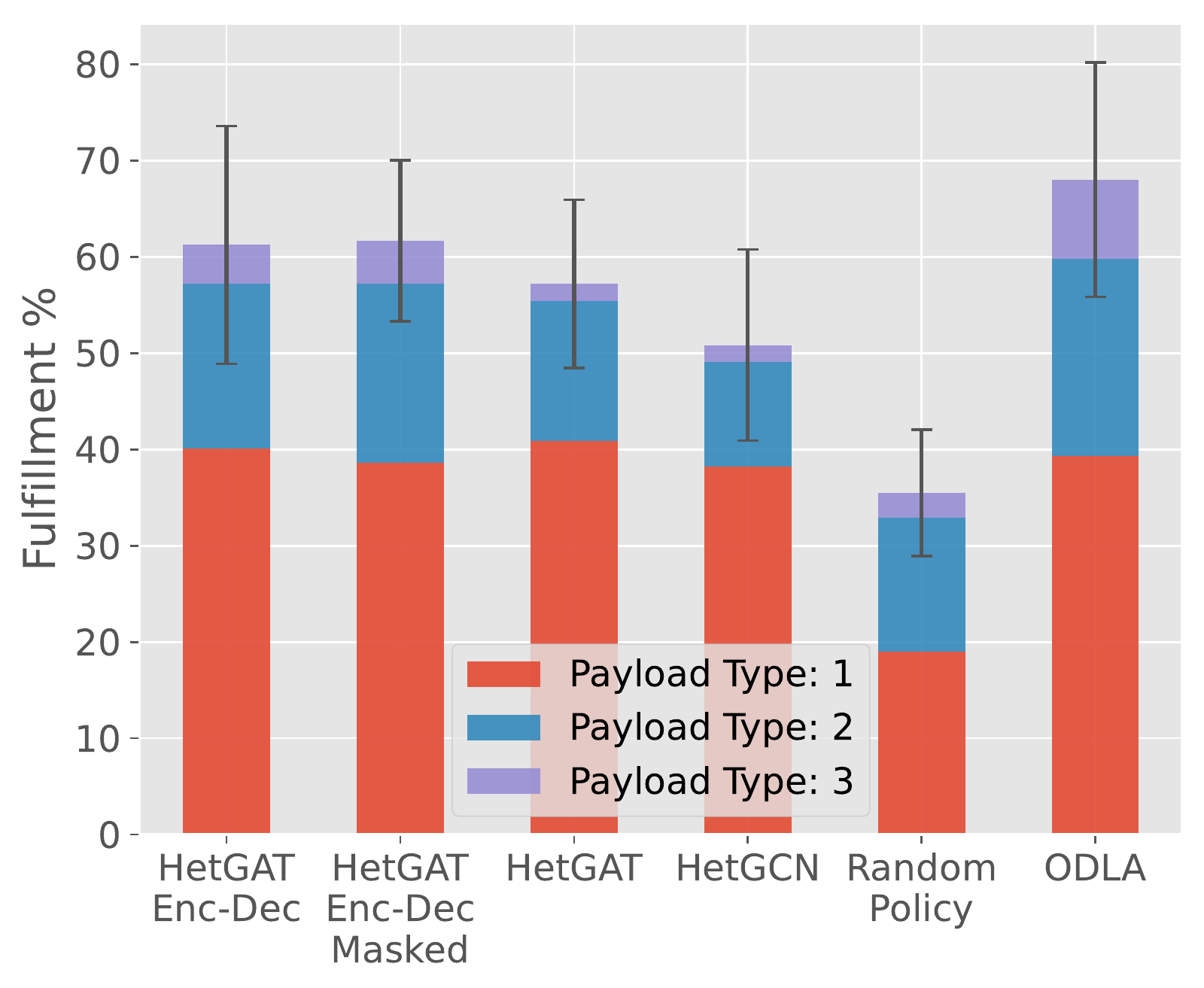}
    \label{fig:eval_fr}}
    \caption{Total fleet reward (a) and fulfillment rates (b) in an one-shot population environment against different policy architectures and ODLA. The environment comprises of 6 UAVs with 2 from each capacity, 10 depots and 12 client nodes.}
    \label{fig:eval}
\end{figure}
We compared the general-sum POSG results to a fully-observed, on-demand payload request assignment (ODLA) approach. 

In ODLA, we assign the requests to the vehicles in a centralized manner by solving a linear assignment problem as discussed in \cite{crouse2016implementing}, minimizing the total cost (\textit{negative} net reward) of the vehicles at a given timestep $t$ where $\mathds{1}_{avail}(v^{t}) = 1$. 
This approach resembles maximizing \ref{eq:obj} at timestep $t=0$ given the full system state iteratively, thus, eliminating the need for having to model the stochastic AAM environment explicitly.
Additionally, it circumvents the requirement to modify the optimization problem as the available number of vehicles changes dynamically.
To solve the optimization problem, we construct a rectangular matrix that tabulates the cost of undertaking each payload by the available vehicles. 
We set the cost to $\infty$ if the payload size exceeds a vehicle's capacity.
From a game-theoretic perspective, ODLA represents an approximate \textit{social optimum} policy. However, we underscore that the centralized assignment nature violates the partial observability, and self-interested constraints imposed on the stochastic game immediately, thus we do not seek a POSG policy that surpasses the socially-optimum policy in the experiments.

Fig. \ref{fig:eval_com}-\ref{fig:eval_fr} show the performances of different policies measured by the total fleet reward and the payload fulfillment percentage in the one-shot population environment.
The fleet composition remains unchanged from that used in the training.
The proposed HetGAT Enc-Dec policy outperforms both HetGAT, HetGCN, and the random policy by a significant margin.
Thanks to the masked policy's ability to explicitly ignore the empty depots in the one-shot partially-observable environment, it performs marginally better (9\% higher fleet rewards, less than 1\% higher delivery fulfillment) compared to the HetGAT Enc-Dec policy.


\subsection{Transferability from One-shot to On-Demand Environments}
\begin{figure}[t]
    \centering
    \subfigure[]
    {\includegraphics[trim={0cm 0cm 0cm 0cm}, clip, width=0.23\textwidth]{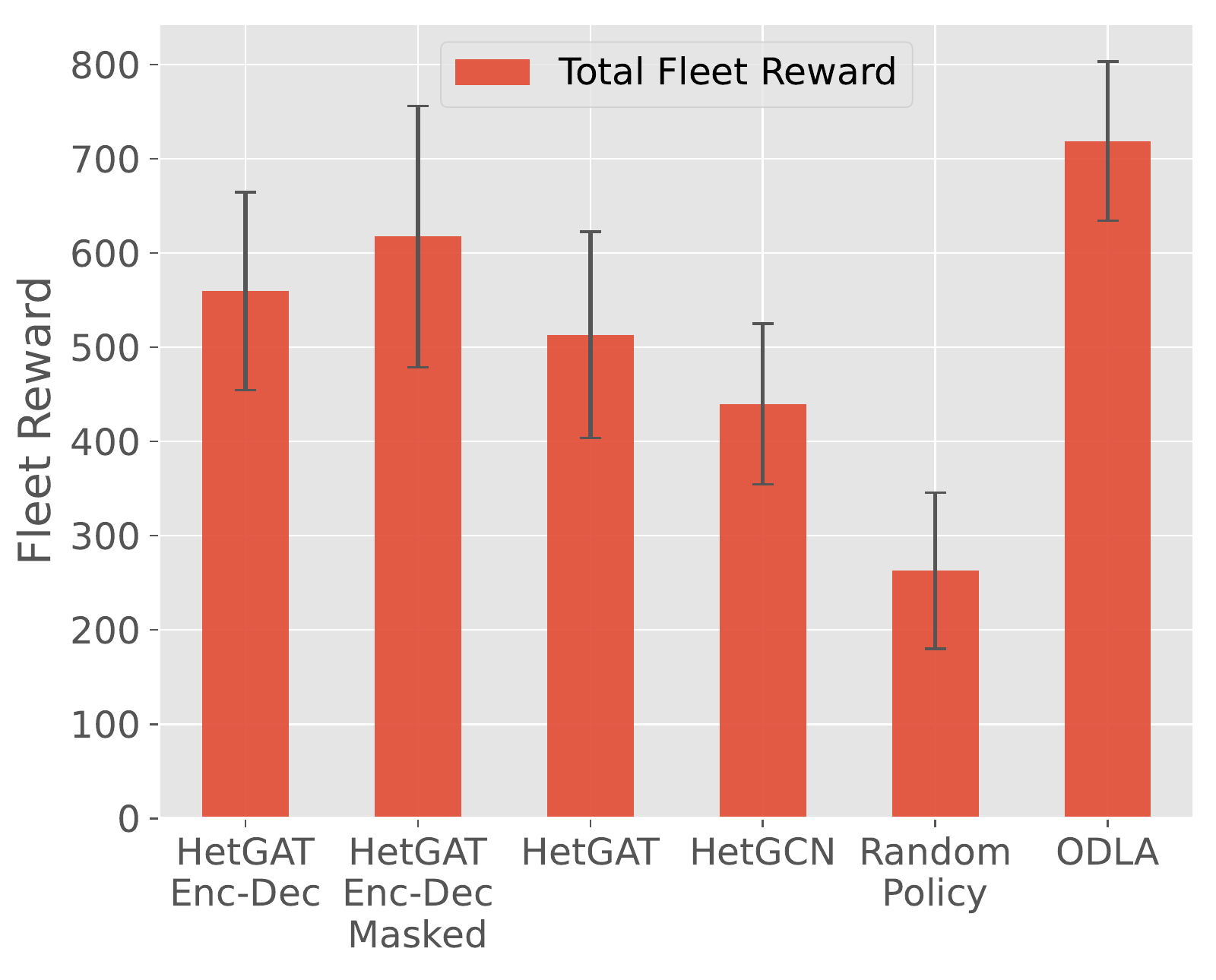}
    \label{fig:od_fr}}
    \subfigure[]
    {\includegraphics[trim={0cm 0cm 0cm 0cm}, clip, width=0.23\textwidth]{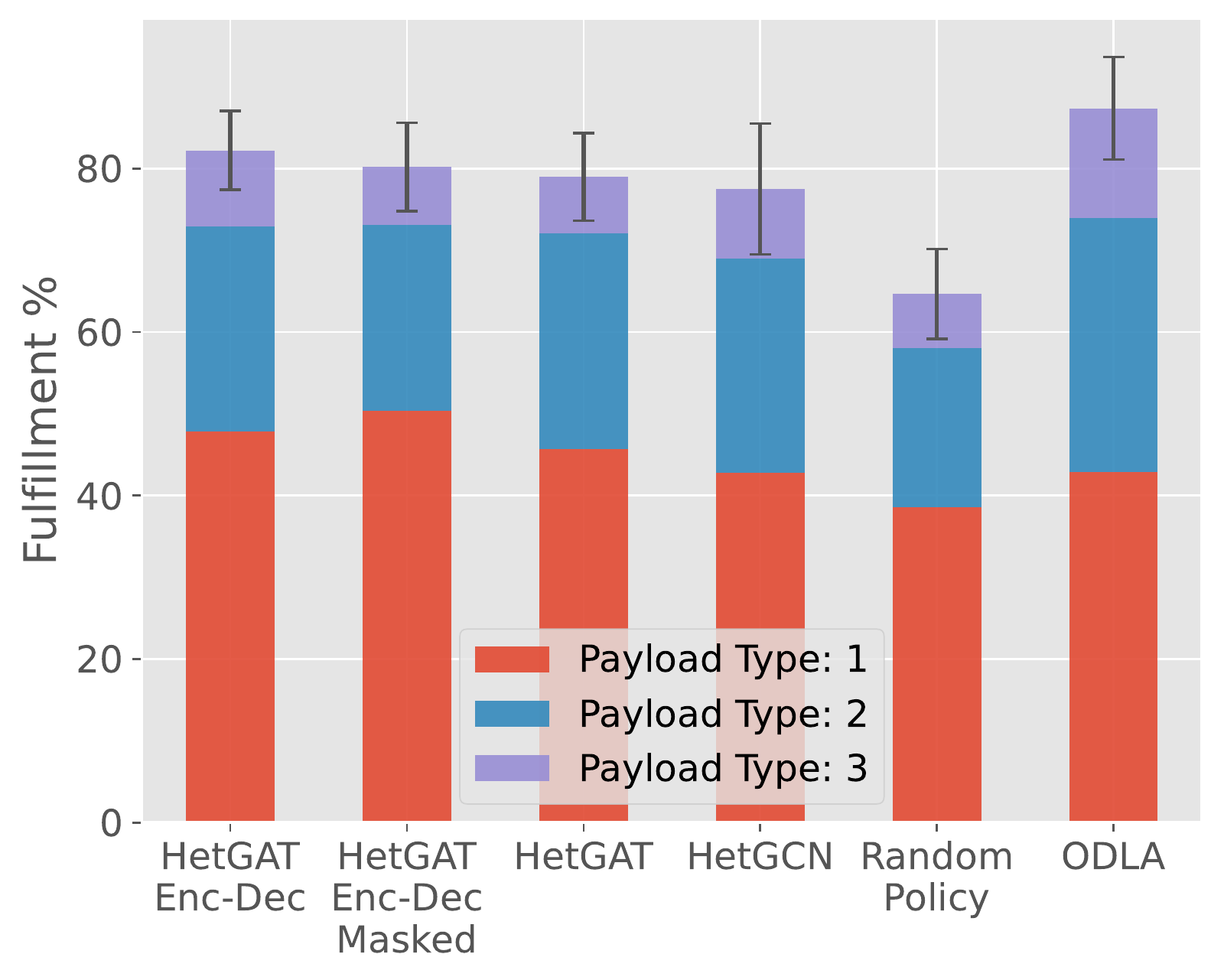}
    \label{fig:od_com}}
    \subfigure[]
    {\includegraphics[trim={0cm 0cm 0cm 0cm}, clip, width=0.23\textwidth]{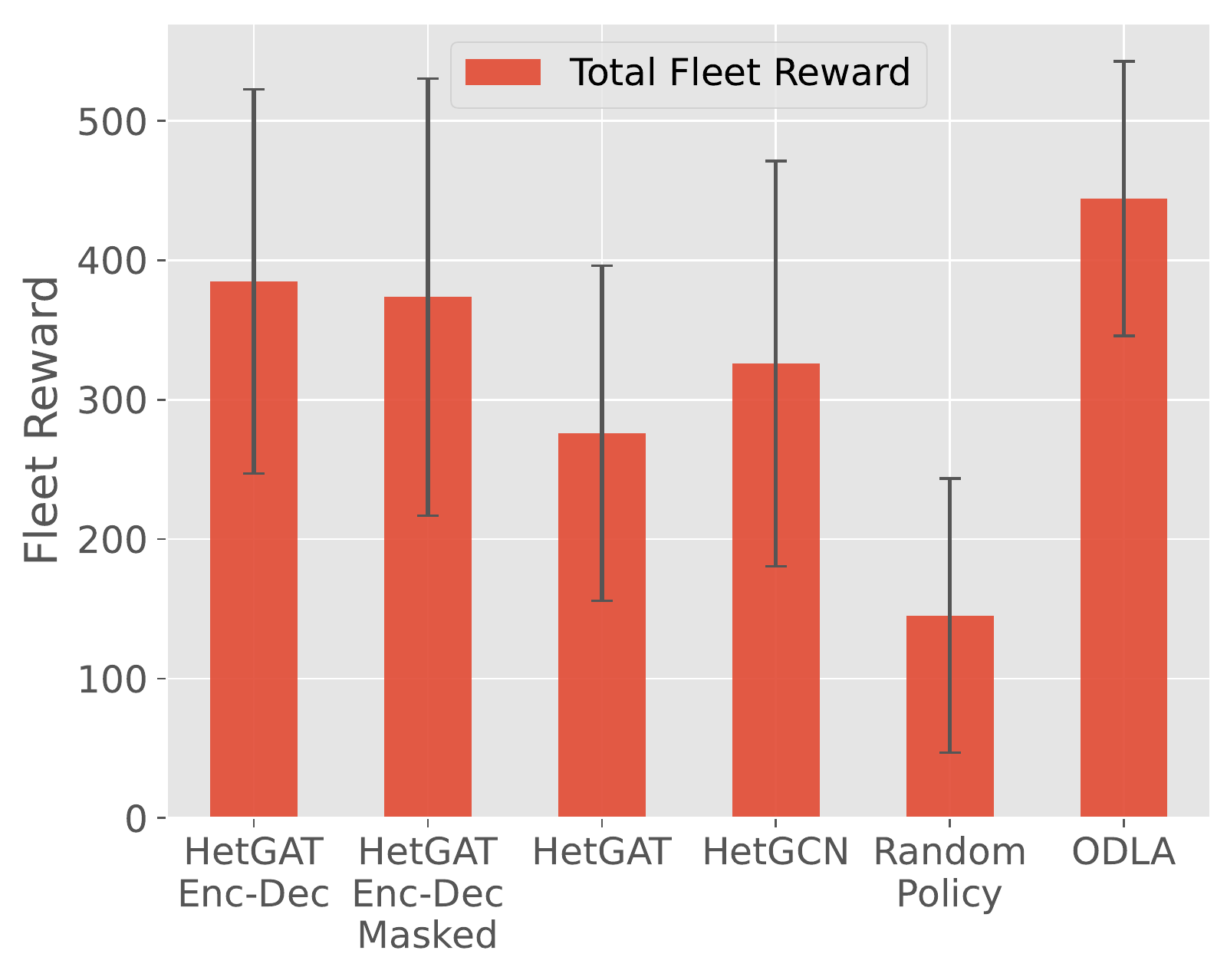}
    \label{fig:scare_fr}}
    \subfigure[]
    {\includegraphics[trim={0cm 0cm 0cm 0cm}, clip, width=0.23\textwidth]{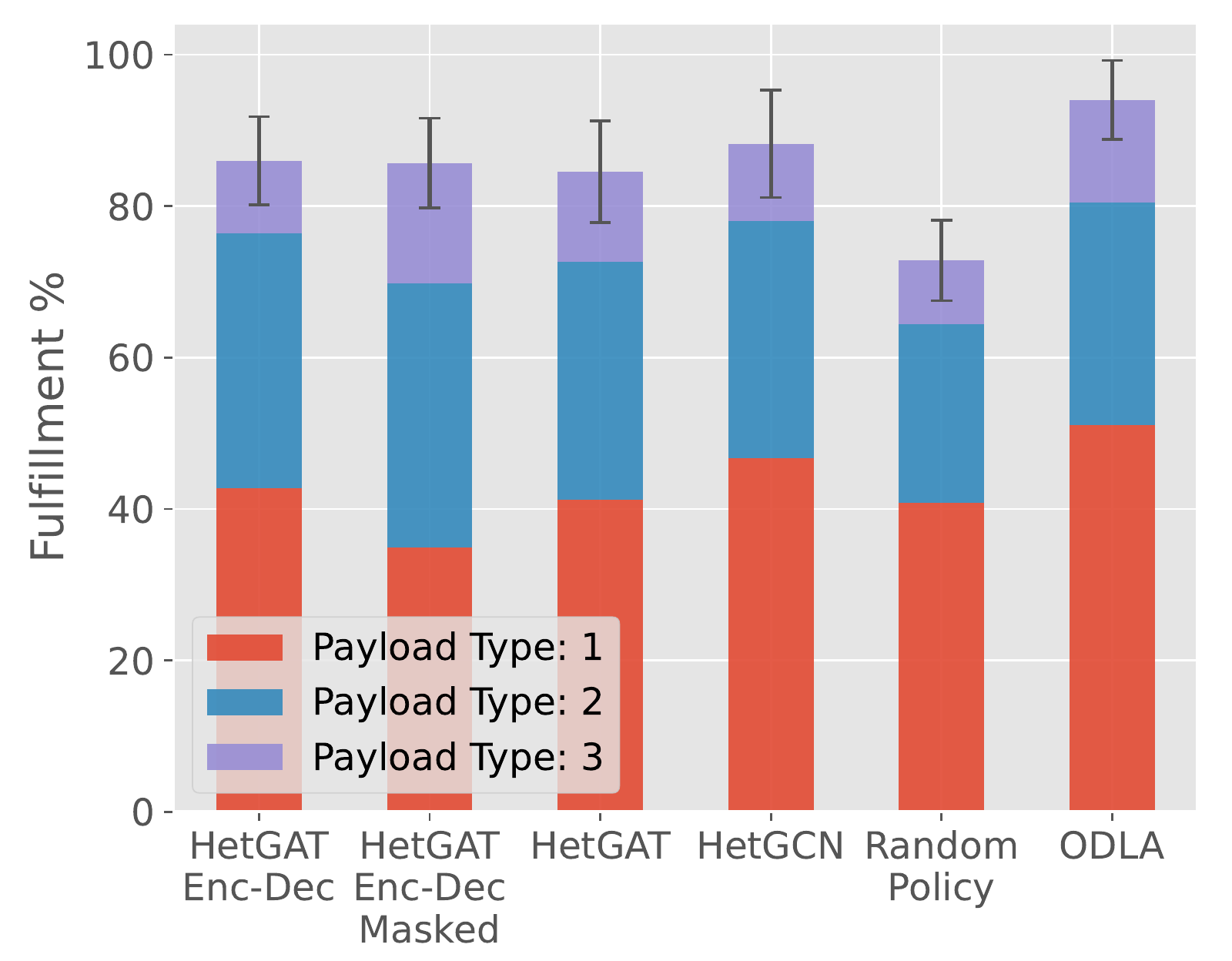}
    \label{fig:scare_com}}
    \caption{Total fleet reward and the percentage of total deliveries fulfilled in high- (\textbf{Top}) and low-yielding (\textbf{Bottom}) on-demand mobility environments against different policy architectures. The two environments received a highest average arrivals of 103.5 and 69.7 payload requests respectively. In the total payload requests arrivals, 49.76\%, 35.62\% and 14.6\% corresponds to category 1, 2 and 3 type payloads. The fleet size remains the same as in the one-shot environment.}
    \label{fig:fleet_reward}
\end{figure}
The real-world mobility environments are often subject to varying demand patterns; i.e., the depots in the same city may experience much lower demands during the after-hours.
To assess the learned policy's ability to handle such demand asymmetries, we experiment with two sets of payload arrival parameters, high- and low-yielding, namely.
The payload requests in the low-yielding scenario use previously unseen, halved Poisson arrival rate parameters $\{0.005, 0.025, 0.0125\}$.
Fig. \ref{fig:od_fr}-\ref{fig:scare_com} report the learned policy's performances in the two environments.

\begin{table*}[]
  \centering
    \caption{Policy generalization to different fleets and on-demand environments for a HetGAT Enc-Dec policy with masking.}
  \begin{tabular}{*{10}{c c c r}}
    \hline
    \hline
      \small
    \textbf{Fleet} & \# \textbf{Depots} & \# \textbf{Clients} & \# \textbf{Payload Req.} & \textbf{\% Fulfillments} & \textbf{Fleet Reward} & \textbf{Rew}. $V_1$ & \textbf{Rew}. $V_2$ & \textbf{Rew}. $V_3$  \\
    \hline
    \hline
    
    \small
    (2,2,2) & 5 & 5 & 51.4$\pm$17.3 & 92.8$\pm$4.9 &372.8$\pm$129.8 & 47.2$\pm$40.7 & 127.3$\pm$53.7 & 198.2$\pm$66.6 \\
    
    \textbf{(2,2,2)} & \textbf{10} & \textbf{12} & 80.1$\pm$5.4 & 80.4$\pm$5.5 & 559.5$\pm$107.8 & 100.7$\pm$69.9 & 207.4$\pm$43.6 & 251.3$\pm$40.3 \\
    
   (0,2,4) & 10 & 12 & 95$\pm$14.2 & 85.5$\pm$6.4 & 695.8$\pm$117.2 & 0$\pm$0 & 210.3$\pm$39.0& 484.2$\pm$89.5 \\
   
    (3,3,4) & 15 & 12 & 141.9$\pm$22.1 & 78.0$\pm$5.3 &999.6$\pm$218.6 & 168.5$\pm$90.4 & 327.6$\pm$75.7& 503.0$\pm$99.1   \\
    
    (3,3,4) & 15 & 24 & 151.6$\pm$19.0 & 81.3$\pm$3.9 &963.9$\pm$163.3 & 149.7$\pm$96.6 & 328$\pm$50.4& 486.2$\pm$62.3  \\
    
    (4,4,7) & 15 & 24 & 254.8$\pm$12.2 & 84.9$\pm$2.4& 2099.6$\pm$136.7 & 377.1$\pm$70.6 & 501.6$\pm$43.4 & 1220.8$\pm$75.0 \\
    \hline
  \end{tabular}
  \label{tab:results}
\end{table*}

The vehicle fleets operating under HetGAT Enc-Dec masked and HetGAT Enc-Dec policies reported the highest fleet rewards in high- and low-yielding scenarios, respectively, compared to the HetGAT, HetGCN, and Random policies.
Although the fleet reward under the masked policy falls marginally behind the unmasked policy in the low-yielding environment, according to Fig. \ref{fig:od_com}, \ref{fig:scare_com}, both the policies report roughly equal fulfillment rates in each environment.
In contrast, the HetGAT and HetGCN -based policies have opted for higher fulfilling ratios at the expense of individual revenue.
HetGCN policy, however, reports higher fleet reward than the HetGAT, and a fulfillment ratio closest to the socially optimal policy in the low-yielding environment.
These statistics show that our HetGAT Enc-Dec Masked can generally achieve the highest individual and fleet rewards, successfully reflecting the self-interest of the agents.
This behavior can greatly benefit high-affinity, commercial AAM, and AMoD fleets, where the vehicles must maximize the owners' revenue while operating under partial observations.

Consistent with the performances in the one-shot environment, the masked HetGAT Enc-Dec reports 9.8\% and roughly 2\% higher fleet rewards and fulfillments compared to the unmasked policy in the on-demand environment with the same rate parameters. 
According to Fig. \ref{fig:eval}, in the one-shot population environment, under the POSG formulation, the HetGAT Enc-Dec policy only falls short 13.8\% and 9\% of the ODLA in fleet reward and the fulfillment ratio, respectively.
Fig. \ref{fig:od_fr}-\ref{fig:od_com} reports similar statistics for the on-demand environment; 14\% and 6\%, respectively.
Following these consistent observations, we confirm that our HetGAT Enc-Dec MARL policies show excellent transferability between one-shot and on-demand payload populations, thus suitable for fleet coordination in either AAM POSG environment without reconfiguration.

\begin{figure}[t]
    \centering
    \subfigure[]
    {\includegraphics[trim={0cm 0cm 0cm 0cm}, clip, width=0.23\textwidth]{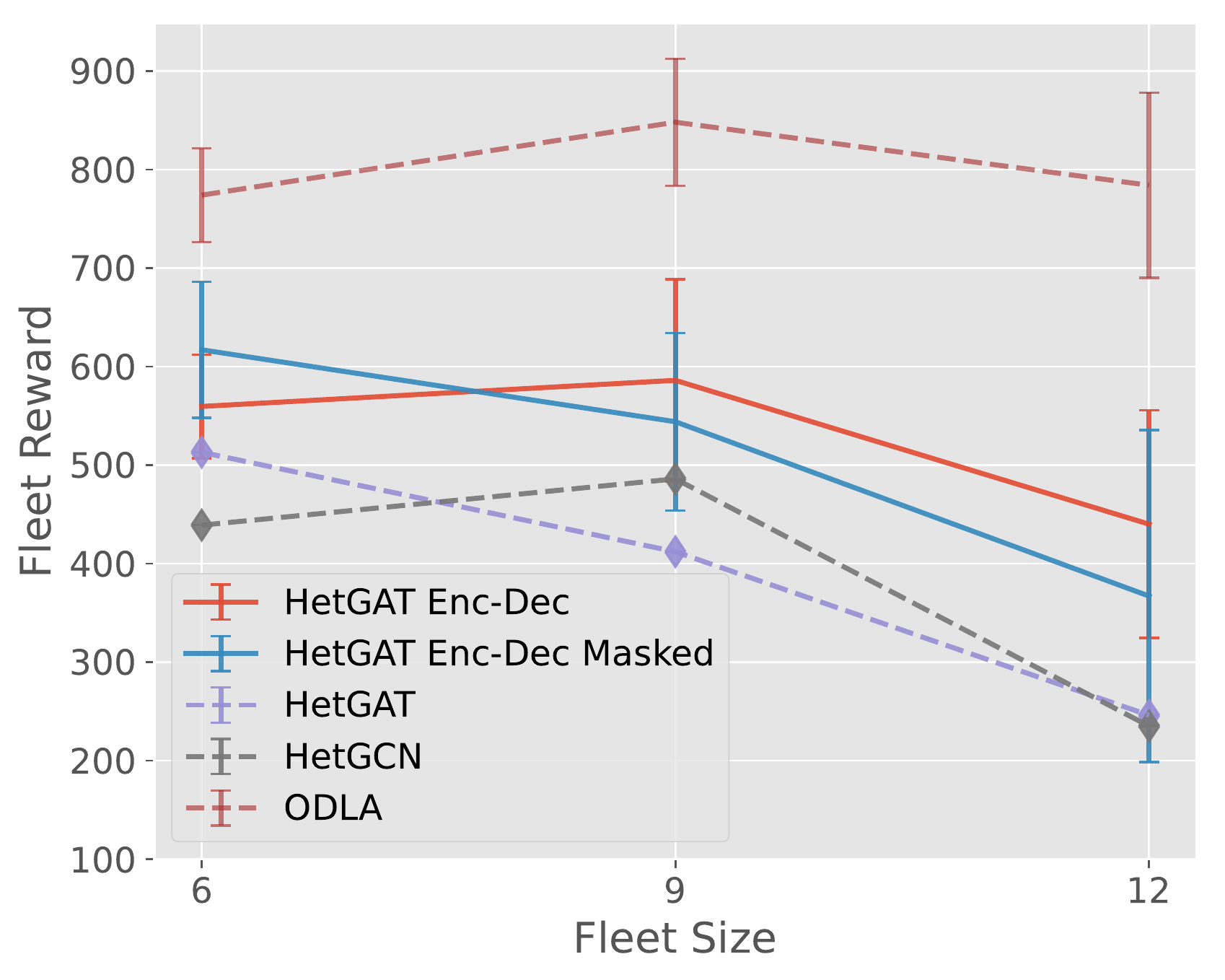}
    \label{fig:d10_rew}}
    \subfigure[]
    {\includegraphics[trim={0cm 0cm 0cm 0cm}, clip, width=0.23\textwidth]{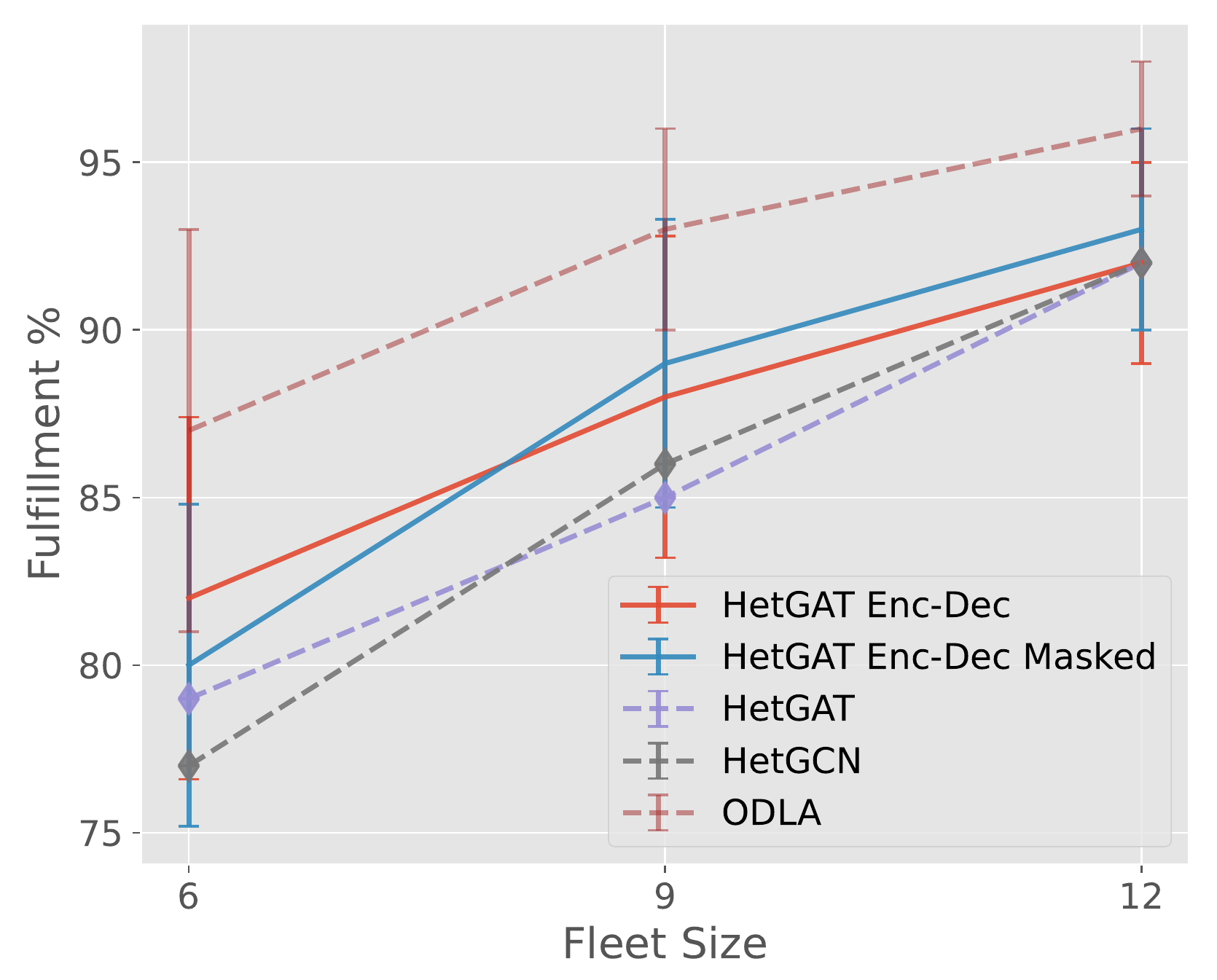}
    \label{fig:d10_com}}
    \subfigure[]
    {\includegraphics[trim={0cm 0cm 0cm 0cm}, clip, width=0.23\textwidth]{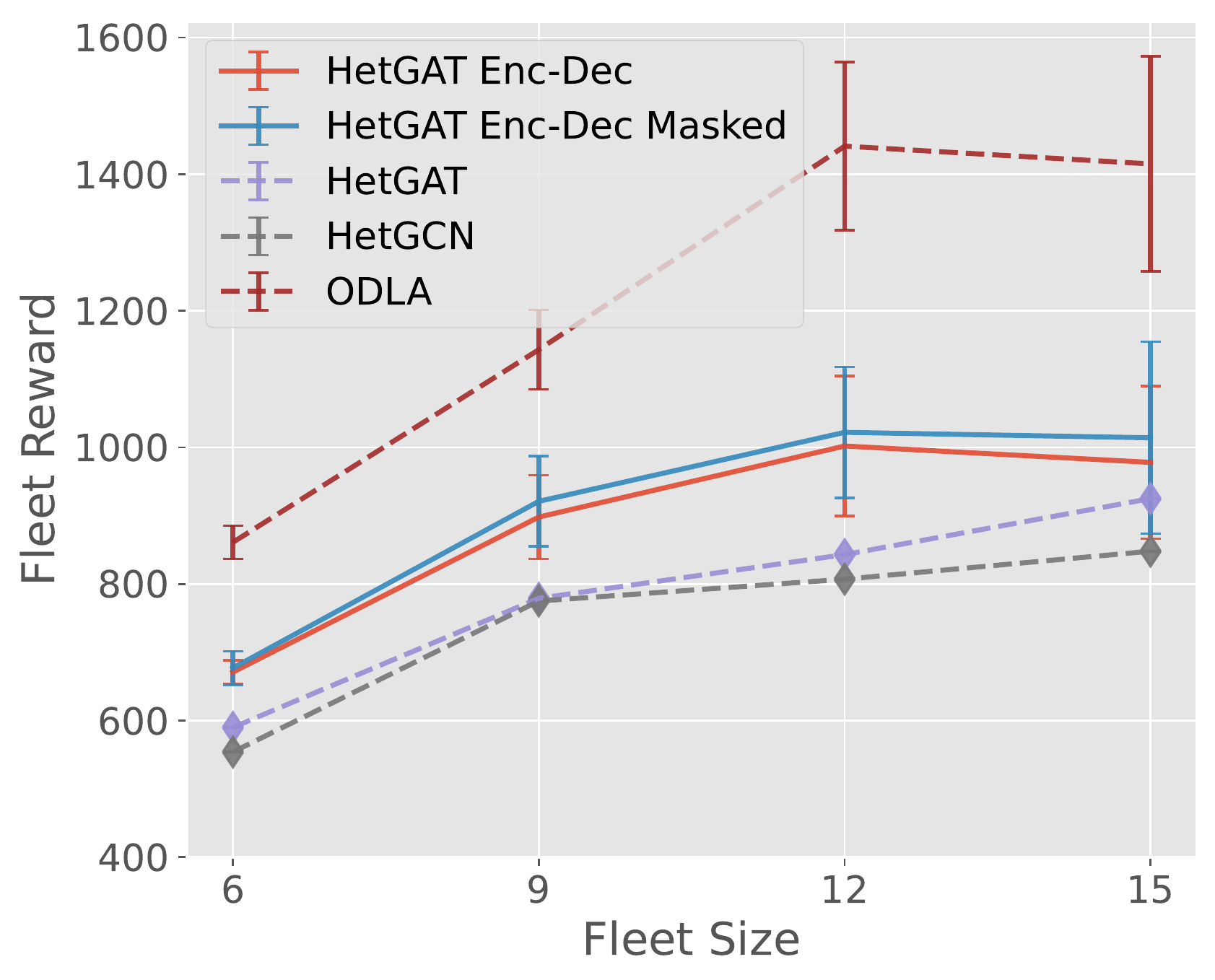}
    \label{fig:d15_rew}}
    \subfigure[]
    {\includegraphics[trim={0cm 0cm 0cm 0cm}, clip, width=0.23\textwidth]{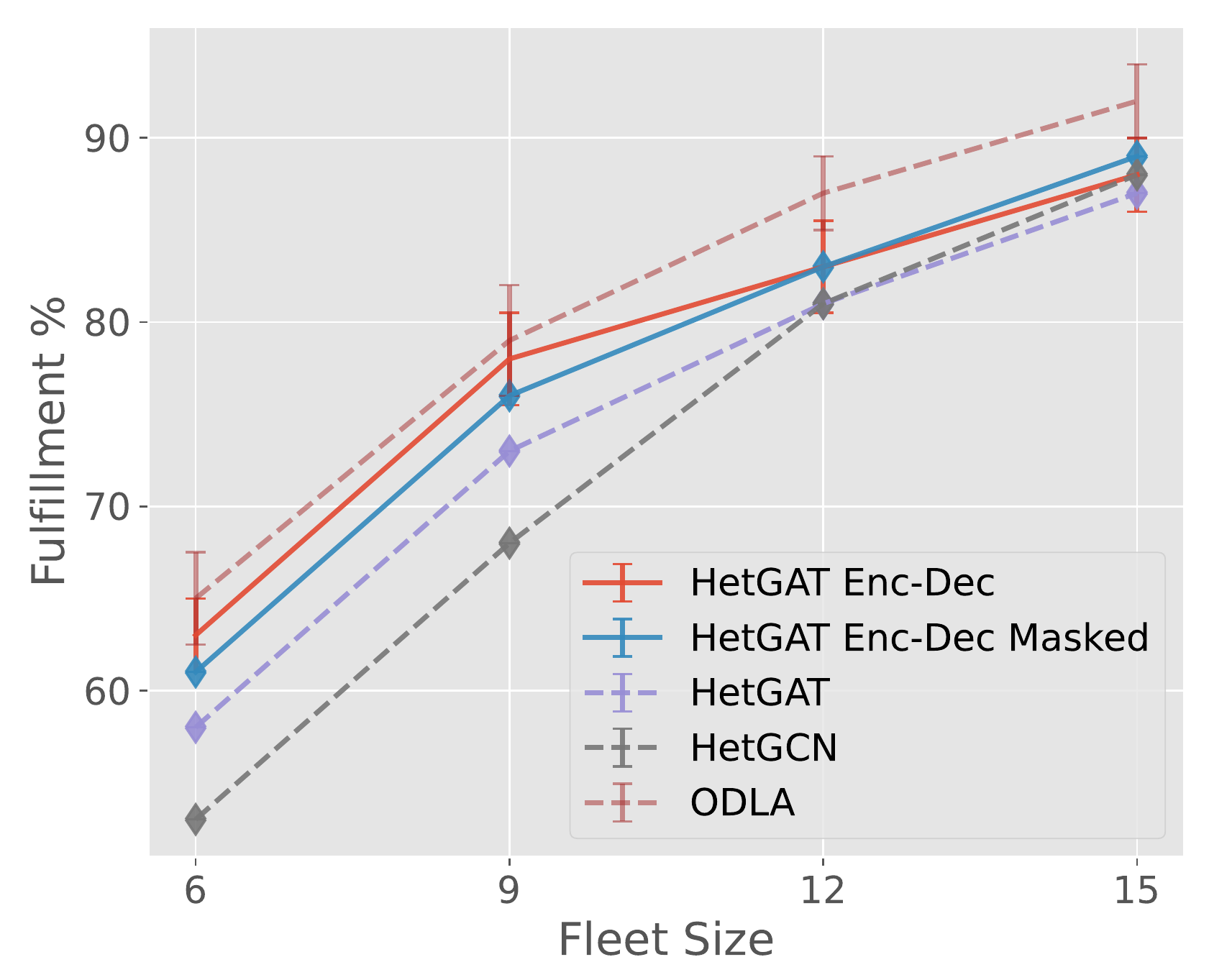}
    \label{fig:d15_com}}
    \caption{Fleet reward and fulfillment rate for different fleet sizes against other generalizable POSG solutions and fully-observable ODLA. \textbf{Top:} An environment with 10 depots, 12 clients with closest $k_d$=5 observation topology. \textbf{Bottom:} An environment with larger service area of 15 depots, 12 clients with closest $k_d$=8 observation topology. Only 50\% error in rewards is showed for HetGAT Enc-Dec and HetGAT policies for clarity.}
    \label{fig:d10}
\end{figure}






\subsection{Generalizability to Varying Fleets and Environments}

To ensure uninterrupted service, and maximize the fleet revenue with the fulfillment rate, a vehicle coordination policy operating in a mobility network must generalize to several fluctuations in the heterogeneous fleet composition, fleet size, and service area. 
We evaluated the performances of our HetGAT Enc-Dec policy to generalize to such fluctuations; by changing the fleet size, vehicle combinations, the number of depots, and the client nodes in the AAM environment.
We report the experimental results in Table \ref{tab:results}.
The ``Fleet" and ``Rew. $V_.$" columns denote 1) the fleet composition as the number of vehicles from each capacity as a tuple 2) and the reward of each vehicle type.
The last row shows the experiment results when there is an increased demand from the clients, a scenario we simulated by doubling the payload arrival rates.

The results show that when increasing the number of depots in the environment while keeping the fleet size constant, all the vehicles receive higher rewards, thus increasing the fleets' collective utility, mainly because a vehicle doesn't need to travel as far to find suitable payloads thanks to the abundance of resources.
Additionally, larger vehicles tend to collect higher rewards than smaller ones due to their ability to attend to more payload types.
Therefore, as one might expect, replacing smaller vehicles with larger ones increase the fleets' reward (\ref{tab:results}, Row 3).
Additionally, by adding more vehicles to the fleet, we can cater to the heightened demand caused by newly added depots.
We experience a slight drop in the fleets' reward when introducing more client nodes to the system who do not contribute with payload requests but only act as the destinations (Recall that the deliveries can happen between any two depot or depot and client nodes).

We account this reduction to the inability of the vehicles to pick up new payloads at their delivery destinations, as opposed to depot-depot deliveries, where a destination may also contain payload requests.
However, when the newly introduced nodes cause a surge of payloads, fleets' collected rewards were observed to increase.
This observation conforms with the real-world notion that service areas with low-demand, scattered destinations are often less preferred by human drivers due to the reduced payoff.

\begin{figure}[t]
    \centering
    \subfigure[]
    {\includegraphics[trim={0cm 0cm 0cm 0cm}, clip, width=0.23\textwidth]{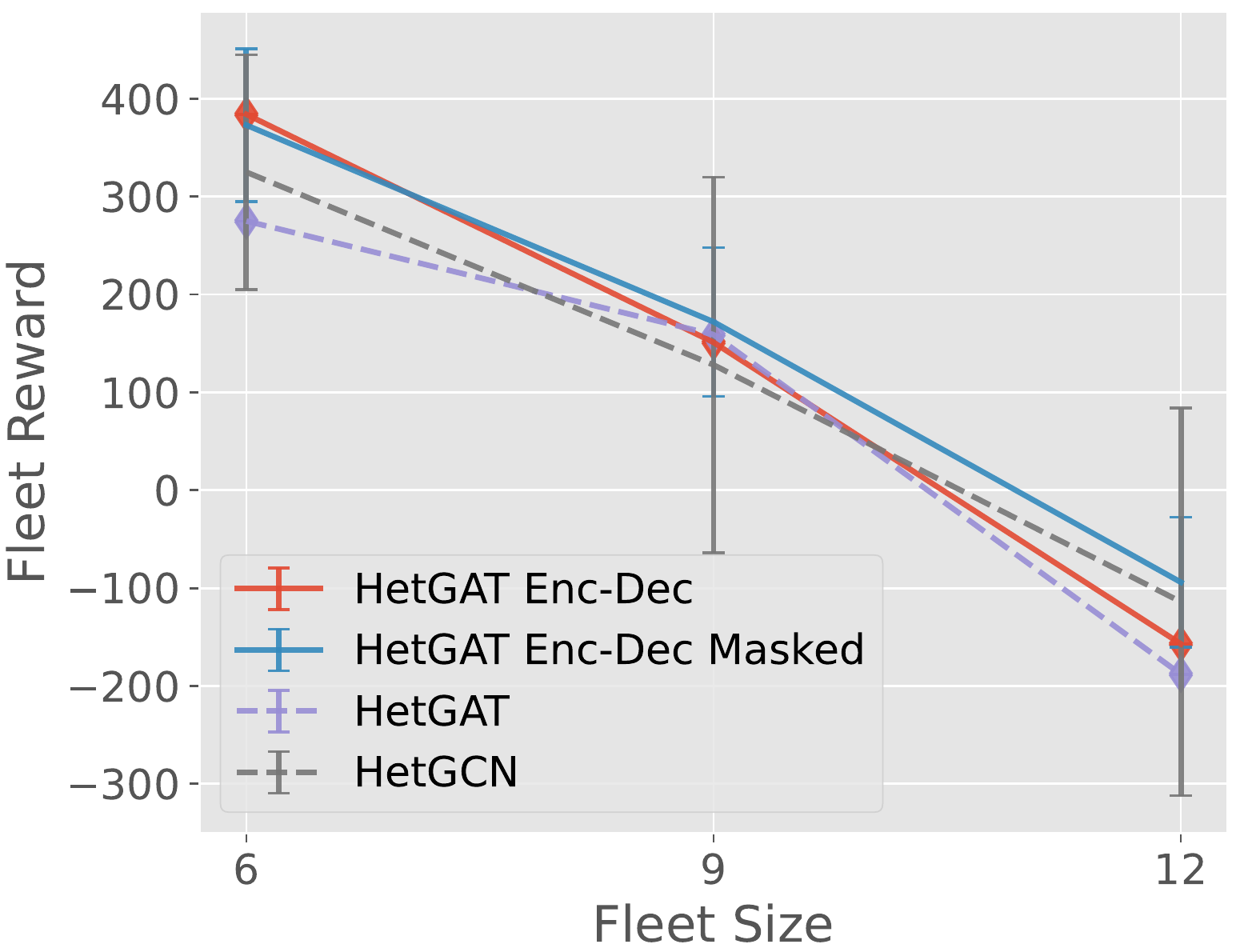}
    \label{fig:ly_kd5_rew}}
    \subfigure[]
    {\includegraphics[trim={0cm 0cm 0cm 0cm}, clip, width=0.23\textwidth]{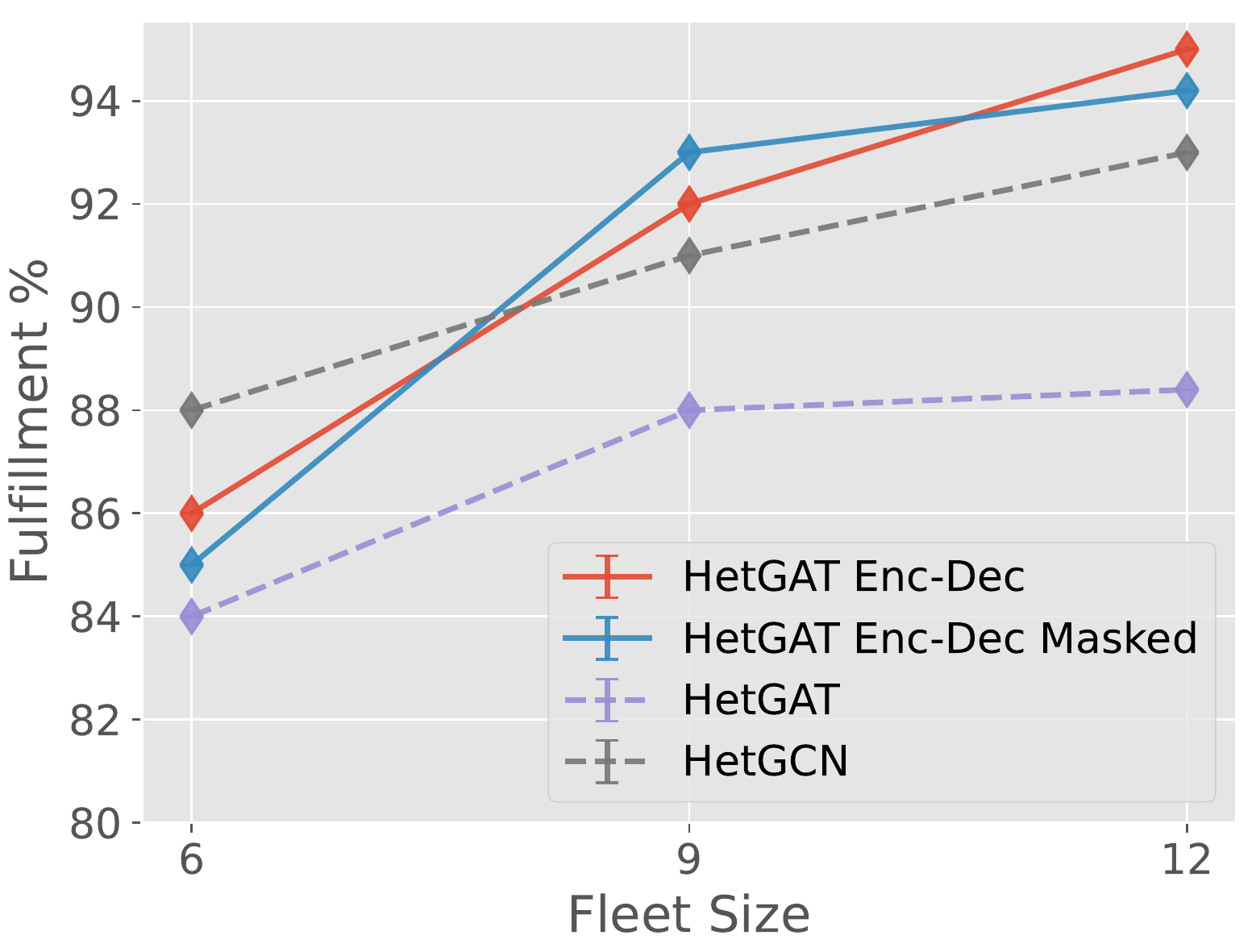}
    \label{fig:ly_kd5_com}}
    
    \subfigure[]
    {\includegraphics[trim={0cm 0cm 0cm 0cm}, clip, width=0.23\textwidth]{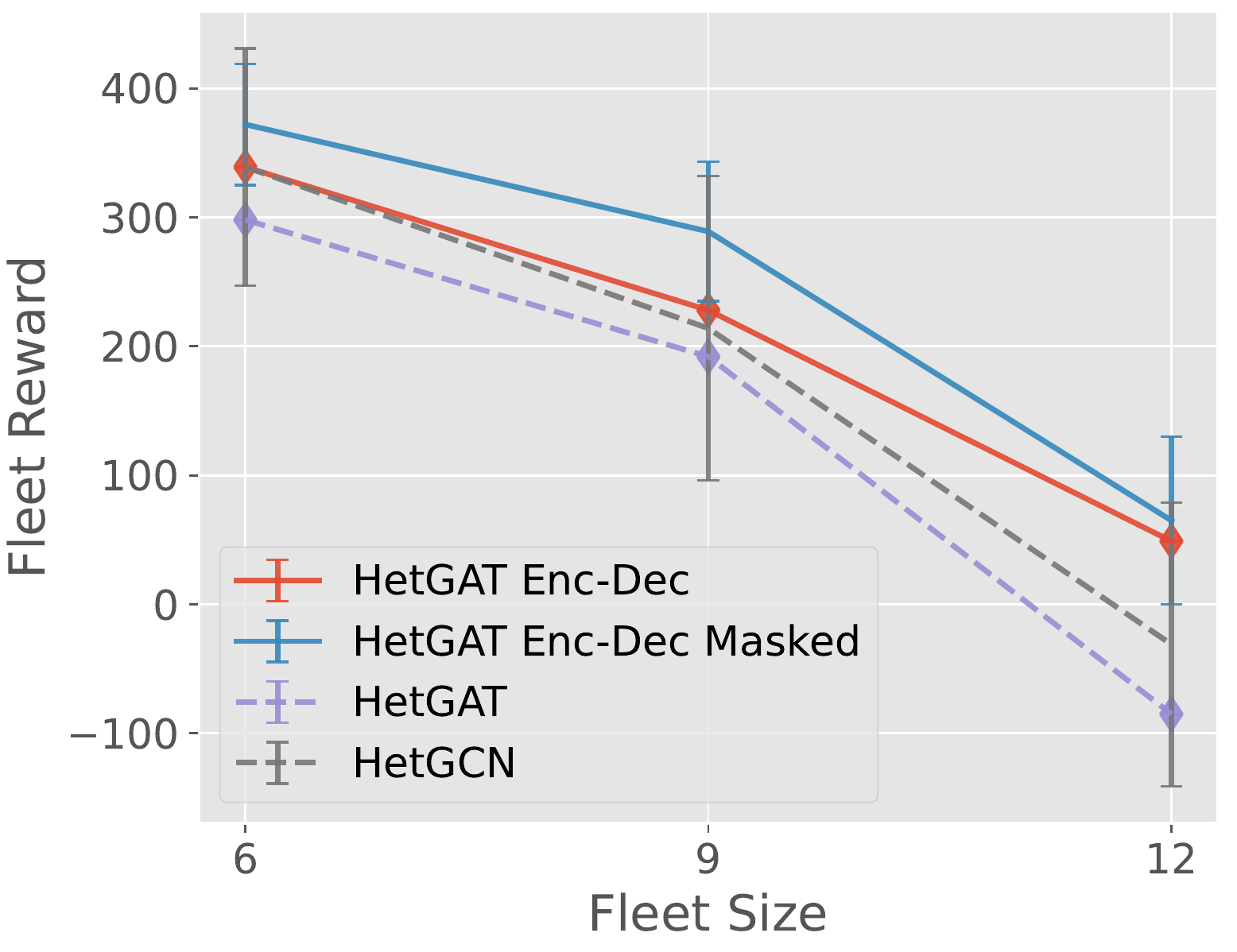}
    \label{fig:ly_kd8_rew}}
    \subfigure[]
    {\includegraphics[trim={0cm 0cm 0cm 0cm}, clip, width=0.23\textwidth]{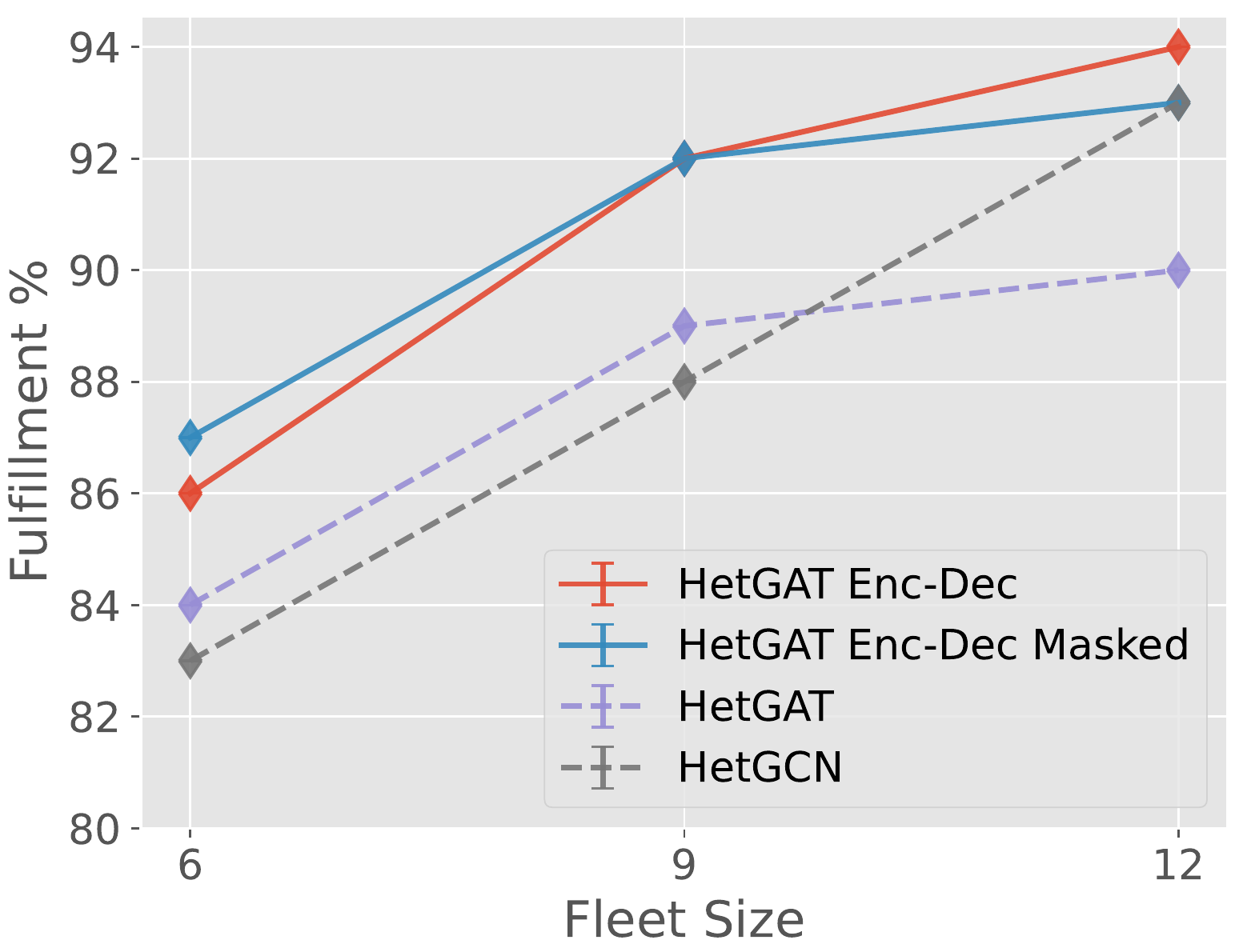}
    \label{fig:ly_kd8_com}}
    \caption{\textbf{Top:} Fulfillment and fleet rewards evaluations for different fleet sizes operating in a low-yielding environment with 10 depots and 12 clients using $k_d$=5 (50\% depots observability). \textbf{Bottom:} Same environment and fleet configurations with a higher observation range $k_d$=8 (80\% depots observability). Only 50\% error in rewards is showed for HetGAT Enc-Dec and HetGCN policies for clarity.}
    \label{fig:d15}
\end{figure}
\begin{figure}[t]
    \centering
    {\includegraphics[trim={0cm 0cm 0cm 0cm}, clip, width=0.44\textwidth]{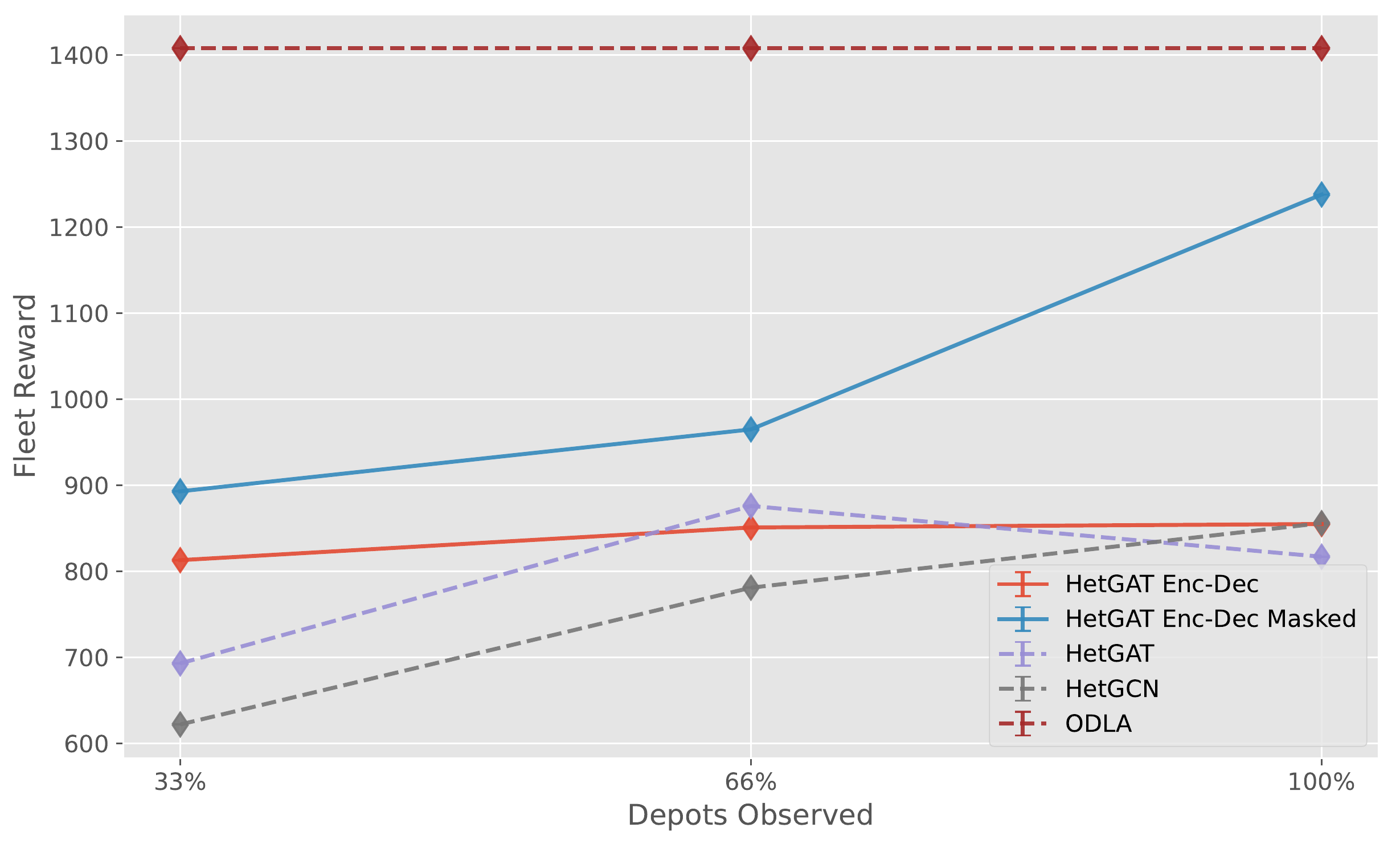}}
    \caption{Fleet reward against the percentage of depots observed. An environment consists of 15 vehicles, 15 depots and 12 clients were used for the experiment. }
    \label{fig:observability}
\end{figure}
\subsection{Policy Generalizability Comparisons}
We evaluated the HetGAT Enc-Dec policy's performances in generalizing to different fleet sizes, service areas, payload arrival imbalances, and observation topologies.
Throughout the experiments, we maintained a 1:1:1 ratio of vehicles from each type in the fleet, $k_v = 5$ vehicle observability, and 50\% depot observation.

Fig. \ref{fig:d10_rew}-\ref{fig:d10_com} show the fleets' collective reward, and the fulfillment rate when changing the fleet size in a service area with 10 depots.
As the fleet size increases we observe a generally downward trend in collective reward acquired by all the policies, due to the increased competition within the fleet.
In other words, when the environment is saturated with vehicles, 1) one's nearby payloads are getting fulfilled sooner, thus causing it to travel farther in sought of suitable payloads, 2) and getting penalized more often from selecting empty depots, thus reducing the net reward.
Fig. \ref{fig:d10_com} shows however, adding new vehicles increases the fulfillment ratio due to the competition.
Fig. \ref{fig:d15_rew}-\ref{fig:d15_com} shows that adding more depots causes the vehicles to obtain higher rewards, and the environment saturates much slower.
The masked HetGAT Enc-Dec achieved the highest collective reward, and fulfillment rate in both environments compared to the other generalizable policy architectures, by only requiring a smaller number of vehicles to saturate the environment. 
In real-world fleets, this characteristic directly translates to lower operational costs, and subsequently higher revenue margins.
We observe that the variance in the fleet reward increases with the fleet size in both the ODLA and POSG solutions, that we account for the inherent stochasticity in the data generation, and the depots' assignment policy $\Psi$.
In other words, we believe that the randomness of generating the delivery requests, and their destinations could accumulate the variance in the fleet reward as the vehicles fulfill more requests. 
Although all the POSG solutions tend to deviate from the social-optimum when exposed to previously unseen fleet combinations, it can be seen that the proposed approaches generalize much better compared to other policy architectures.


We compared the agents' performances in low-yielding environments by keeping the fleet composition unchanged.
Fig. \ref{fig:ly_kd5_rew} shows that introducing more agents to resource-limited environments further degrades the fleet reward in all four generalizable policy models.
Fig. \ref{fig:ly_kd8_rew} shows that the fleet rewards under both HetGAT Enc-Dec policies degrade more gracefully than other GNN policies while achieving the highest fulfillment rates in the low-yielding environments.
Interestingly, Fig. \ref{fig:ly_kd8_rew}-\ref{fig:ly_kd8_com} shows that changing the vehicles' observation range up to 80\% of the available closest depots to result in improved performances of the HetGAT Enc-Dec policies.
We state that this behavior highlights the ability of the HetGAT Enc-Dec policy's ability to incorporate new information to improve the quality of the decisions.
We also notice that despite not using the attention mechanism, HetGCN to outperform HetGAT in scalability experiments, as shown in Fig. \ref{fig:ly_kd5_com} - \ref{fig:ly_kd8_com}. 

\subsection{Adaptability to Varying Observation Topologies}
\begin{figure}[t]
    \centering
    {\includegraphics[trim={0cm 0cm 0cm 0cm}, clip, width=0.4\textwidth]{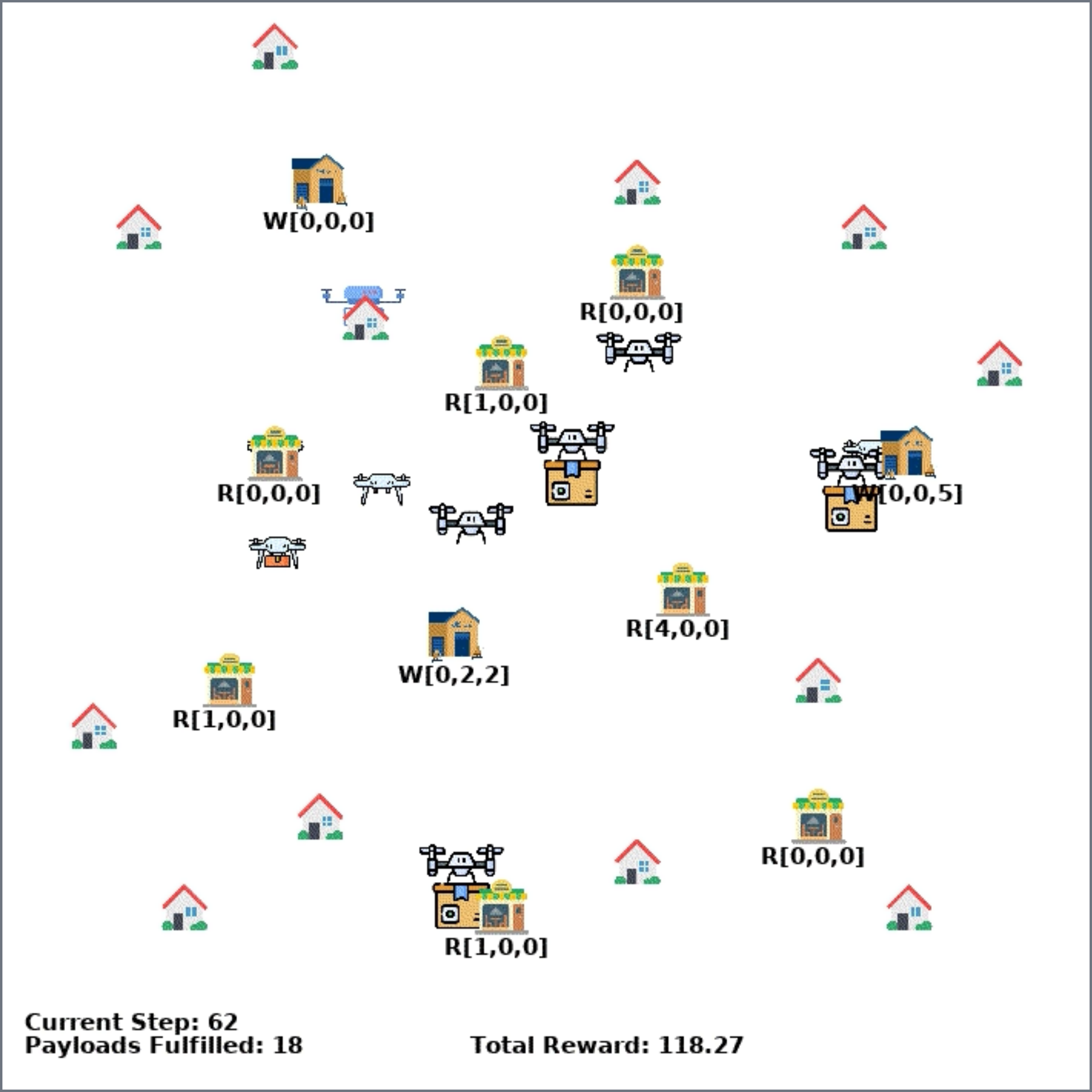}}
    \caption{A snapshot of a densely populated simulation environment used for experiments that consists of 10 vehicles, 10 depots, and 12 clients. The environment contains 3 types of vehicles with different capacities and 2 types of depots (R, W) with different expected payload capacities. The number of available payloads from each type is denoted below the depots..}
    \label{fig:ss_full}
\end{figure}

We evaluate the fleets' reward and the fulfillment rate against different observation topologies.
Throughout the experiment, we kept the number of observed vehicles fixed while increasing the visibility of the depots: a realistic consideration as disclosing the other vehicle's locations is less desirable in the pursuit of higher rewards due to privacy concerns.
Fig. \ref{fig:observability} shows that our masked HetGAT Enc-Dec policy increases the fleets' reward exponentially as the observability reaches 100\%, a contrasting difference to the other policies.
This showcases our approach's ability to handle time-varying observational topologies, which often arise in AAM due to the stochasticity in wireless networks. 
Briefly, to maximize the agents' rewards in low-yielding environments, we advocate 1) operating the vehicle agents under the masked HetGAT Enc-Dec policy, and 2) revealing more depot information to the agents.
All the generalizable POSG solutions presented herewith tend to deviate from the social optimum as the fleet size increases, especially under low-yielding environments, see Fig. \ref{fig:ly_kd5_rew}-\ref{fig:ly_kd8_rew}. 
However, from Fig. \ref{fig:observability} it can be seen that the HetGAT Enc-Dec approach we have proposed gets closer to the social optimum when exposed to more observations on the environment when coupled with a rebalancing mask.
Fig. \ref{fig:ss_full} shows a densely populated simulation environment used in the experiments.


\section{Discussion and Conclusion}
We present a novel, generalizable, multi-agent fleet autonomy for coordinating heterogeneous mobility fleets in a decentralized manner under partial observations building on HetGAT and encoder-decoder neural networks.
Extensive experiments conducted under different fleet combinations, service areas, observational topologies, and fulfillment request arrival rates showed that agents fleets operating under HetGAT Enc-Dec policies outperform the other generalizable policy architectures.
The novel fleet rebalancing mask further improved the ability of our method to perform in low-yielding on-demand mobility networks and especially to incorporate the observational topologies beyond that were used in the training time into the decision-making.
\rebutt{The new insights show that the proposed HetGAT Enc-Dec, when coupled with an agent rebalancing mask could yield more close-to-optimal results.}
The two policy architectures we proposed further achieved the highest fleet reward using the minimum number of vehicles compared to other generalizable POSG solutions while maximizing the fulfillment ratios: a highly sought-after characteristic for commercial mobility fleets.

\bibliographystyle{ieeetr}

\bibliography{root}




\appendices
\vspace{10pt}
\section{Neural Network Architectures}
For the fully-connected neural network $\mathrm{fc\_val}$, we used a two hidden-layers with the input features vectorized together. 
The output layers in the $\mathrm{fc\_val}$ maps from the number of depots to a scalar value.

\begin{table}[htbp!]
\centering
\caption{Training Parameters}
\begin{tabular}{|l|c|c|c|}
\hline
\multicolumn{1}{|c|}{\textbf{Model/Training Params.}} &
\multicolumn{1}{c|}{\textbf{Batch Size}} &
  \multicolumn{1}{c|}{\begin{tabular}[c]{@{}c@{}}\textbf{Minibatch}\\ \textbf{Size}\end{tabular}} &
  \multicolumn{1}{c|}{\begin{tabular}[c]{@{}c@{}}\textbf{Entropy}\\ \textbf{Coeff.}\end{tabular}}
  \\ \hline
HetGAT Enc-Dec & 1200 & 48 & $10^{-2}$ \\ \hline
HetGAT/HetGCN  & 1200 & 48 & $10^{-3}$  \\ \hline
\end{tabular}
\label{tab:training}
\vspace{-2em}
\end{table}

\vspace{10pt}

\label{app:nn}

\begin{table}[htbp!]
\centering
\caption{Different HetGAT neural network architectures.}
\begin{tabular}{|ll|c|c|c|}
\hline
\multicolumn{2}{|l|}{\textbf{Layers/Type}}                                  & \begin{tabular}[c]{@{}l@{}}HetGAT-Enc\\ ($\mathcal{V}$, $\mathcal{D}$, $\mathcal{P}$)\end{tabular}          & \begin{tabular}[c]{@{}l@{}}HetGAT-Dec\\ ($\mathbf{g}$, $\mathcal{D}$, $\mathbf{val}$)\end{tabular}           & \begin{tabular}[c]{@{}l@{}}HetGAT \\ HetGCN\\ ($\mathcal{V}$, $\mathcal{D}$, $\mathcal{P}$)\end{tabular}              \\ \hline
\multicolumn{1}{|l|}{\multirow{2}{*}{\textbf{Layer 1}}} & \begin{tabular}[c]{@{}l@{}}\textbf{Input/Output}\\ \textbf{Dim.}\end{tabular} & \begin{tabular}[c]{@{}c@{}}(5,4,4)\\ (32,32,32)\end{tabular}    & \begin{tabular}[c]{@{}c@{}}(192,64,32)\\ (48,48,48)\end{tabular} & \begin{tabular}[c]{@{}c@{}}(5,4,4)\\ (32,32,32)\end{tabular}    \\ \cline{2-5} 
\multicolumn{1}{|l|}{}                         & \textbf{Att. Heads}                                                  & 8                                                               & 8                                                                & 8                                                               \\ \hline
\multicolumn{1}{|l|}{\multirow{2}{*}{\textbf{Layer 2}}} & \begin{tabular}[c]{@{}l@{}}\textbf{Input/Output}\\ \textbf{Dim.}\end{tabular} & \begin{tabular}[c]{@{}c@{}}(32,32,32)\\ (32,32,32)\end{tabular} & \begin{tabular}[c]{@{}c@{}}(48,48,48)\\ (64,64,64)\end{tabular}  & \begin{tabular}[c]{@{}c@{}}(32,32,32)\\ (32,32,32)\end{tabular} \\ \cline{2-5} 
\multicolumn{1}{|l|}{}                         & \textbf{Att. Heads}                                                  & 8                                                               & 1                                                                & 8                                                               \\ \hline
\multicolumn{1}{|l|}{\multirow{2}{*}{\textbf{Layer 3}}} & \begin{tabular}[c]{@{}l@{}}\textbf{Input/Output}\\ \textbf{Dim.}\end{tabular} & \begin{tabular}[c]{@{}c@{}}(32,32,32)\\ (64,64,64)\end{tabular} & \begin{tabular}[c]{@{}c@{}}fc\_val (64)\\ 1\end{tabular}         & \begin{tabular}[c]{@{}c@{}}(32,32,32)\\ (64,1,64)\end{tabular}  \\ \cline{2-5} 
\multicolumn{1}{|l|}{}                         & \textbf{Att. Heads}                                                 & 1                                                               & 1                                                                 & 1                                                               \\ \hline
\end{tabular}
 \label{tab:nn}
\end{table}

We show the implementation details for each neural network used in this work in Table \ref{tab:nn}. 
The HetGCN network uses the same combination of layers as the HetGAT, except for the attention heads, the distinguishing feature of attention type neural networks.
The HetGAT network we used for the experiments shares the same architecture as the HetGAT-Encoder.
For the training, we used a learning rate that decayed over a course of $300000$ active timesteps from $10^{-4}$ to $10^{-5}$. 
We list all the training parameters used for PPO in Table \ref{tab:nn} and below.

\begin{itemize}
    \item SGD iterations - 8
    \item Value function loss coefficient - $5\times10^{-3}$
    \item Clip parameter - 0.1
    \item $\lambda$ (PPO) - 0.95
    \item $\gamma$ - 0.99
\end{itemize}

\end{document}